\PassOptionsToPackage{table,dvipsnames}{xcolor}
\documentclass[sigconf,nonacm]{acmart}

\usepackage{amssymb}
\usepackage{array}
\usepackage{float}
\usepackage{fvextra}
\fvset{breaksymbolleft={},breaksymbolright={}}
\usepackage[most]{tcolorbox}
\usepackage{subcaption}
\usepackage{xspace}
\usepackage{array}
\usepackage{tabularx}
\usepackage{multicol}
\usepackage{tikz}
\usetikzlibrary{positioning,arrows.meta,shapes.geometric,calc,fit,backgrounds}
\usepackage{pgfplots}
\pgfplotsset{compat=1.18}

\newcommand{\arena}{\mbox{LLM-SoccerArena}\xspace}
\newcommand{\arenalink}{\href{https://llm-soccerarena.com}{\texttt{\small\color{blue}https://llm-soccerarena.com}}}

\newcommand{\googleicon}{\raisebox{-0.2ex}{\includegraphics[height=1.7ex,keepaspectratio]{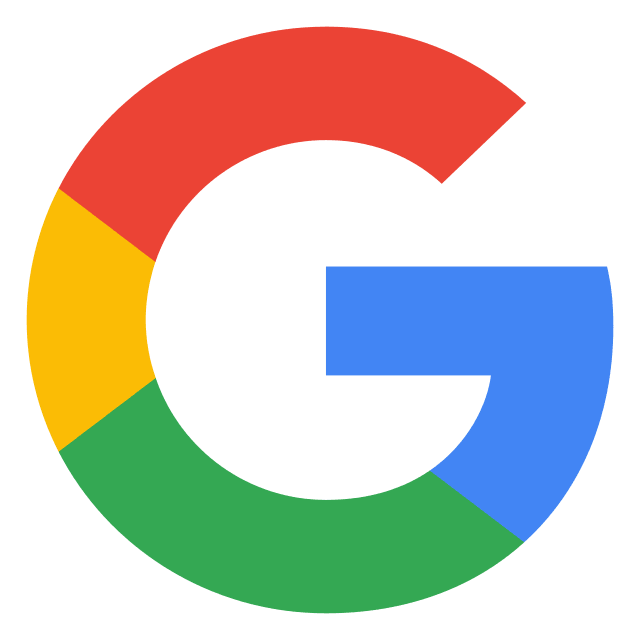}}\hspace{0.35em}}
\newcommand{\openaiicon}{\raisebox{-0.2ex}{\includegraphics[height=1.7ex,keepaspectratio]{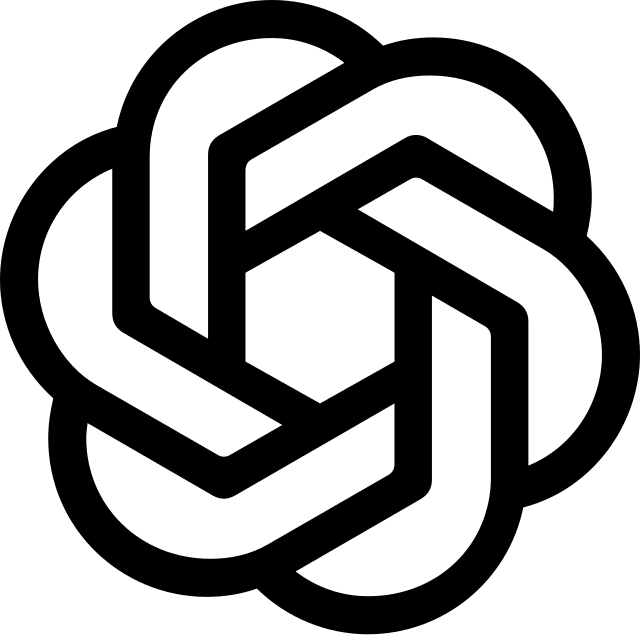}}\hspace{0.35em}}
\newcommand{\deepseekicon}{\raisebox{-0.2ex}{\includegraphics[height=1.7ex,keepaspectratio]{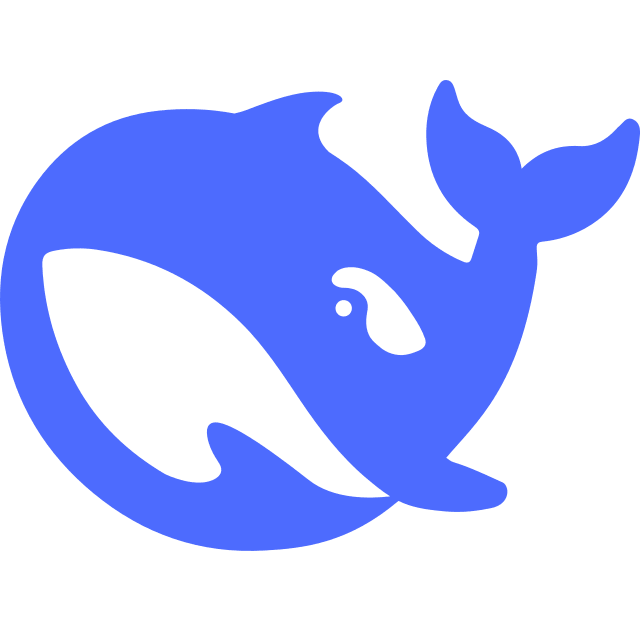}}\hspace{0.35em}}
\newcommand{\xaiicon}{\raisebox{-0.2ex}{\includegraphics[height=1.7ex,keepaspectratio]{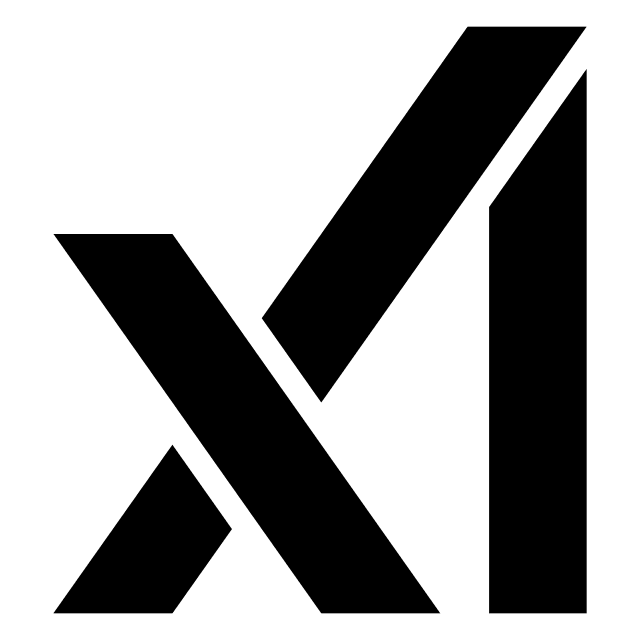}}\hspace{0.35em}}
\newcommand{\alibabaicon}{\raisebox{-0.2ex}{\includegraphics[height=1.7ex,keepaspectratio]{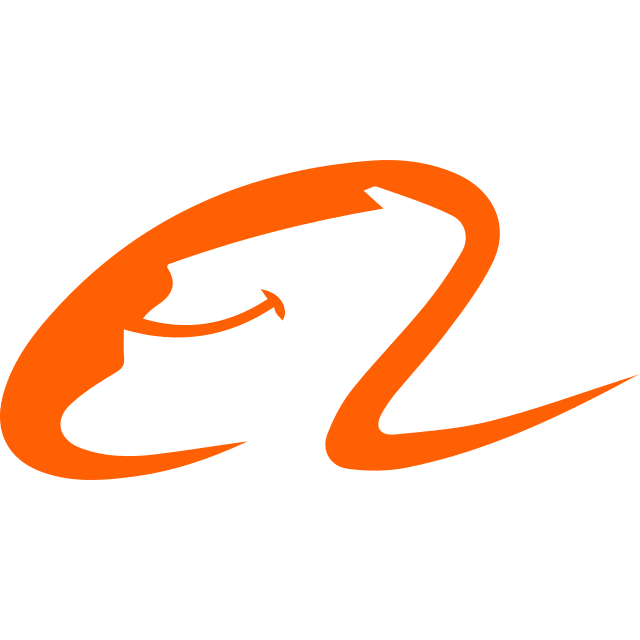}}\hspace{0.35em}}
\newcommand{\anthropicicon}{\raisebox{-0.2ex}{\includegraphics[height=1.7ex,keepaspectratio]{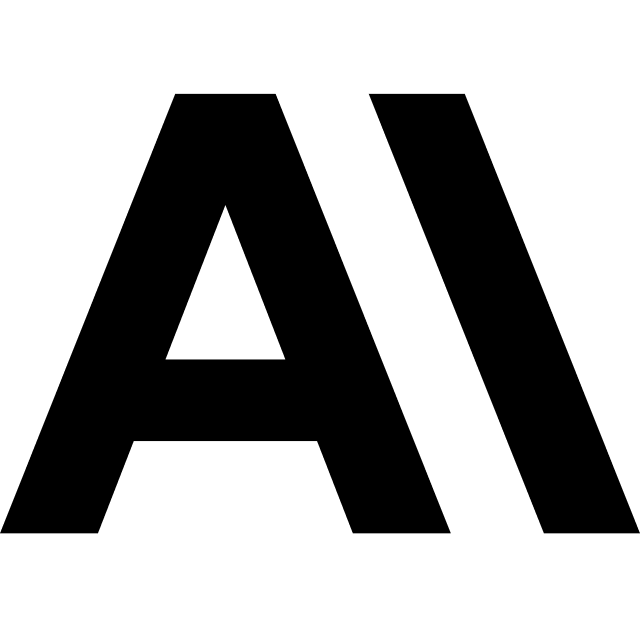}}\hspace{0.35em}}
\newcommand{\mistralicon}{\raisebox{-0.2ex}{\includegraphics[height=1.7ex,keepaspectratio]{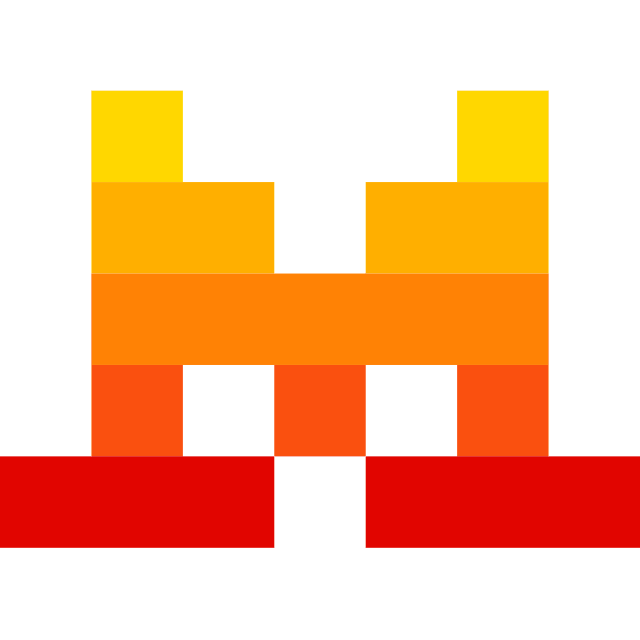}}\hspace{0.35em}}
\newcommand{\soccerballicon}{\raisebox{-0.2ex}{\includegraphics[height=1.7ex,keepaspectratio]{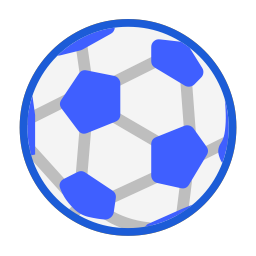}}}
\newcommand{\trophyicon}{\raisebox{-0.2ex}{\includegraphics[height=1.7ex,keepaspectratio]{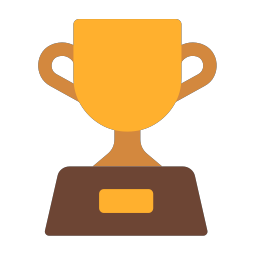}}}
\newcommand{\clipboardicon}{\raisebox{-0.2ex}{\includegraphics[height=1.7ex,keepaspectratio]{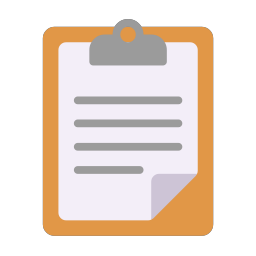}}}
\newcommand{\githubicon}{\raisebox{-0.2ex}{\includegraphics[height=1.7ex,keepaspectratio]{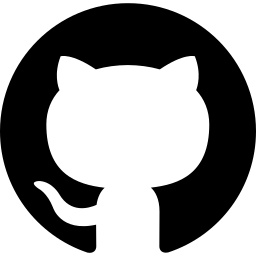}}}

\definecolor{draftblue}{RGB}{34,91,145}
\definecolor{draftlight}{RGB}{239,246,252}
\definecolor{todoorange}{RGB}{176,91,0}
\definecolor{todolight}{RGB}{255,246,230}

\usepackage{pifont}
\newcommand{\yes}{\textcolor{green!55!black}{\ding{51}}}
\newcommand{\partly}{\textcolor{orange!85!black}{\textbf{\textasciitilde}}}
\newcommand{\no}{\textcolor{BrickRed}{\ding{55}}}

\usepackage{tikz}

\definecolor{tikzblue}{HTML}{082567}   
\definecolor{tikzgreen}{HTML}{145214} 
\definecolor{tikzred}{HTML}{6E0000}    

\newcommand{\circnum}[2][tikzblue]{%
\tikz[baseline=-0.55ex]{
\node[
    circle,
    draw=#1,
    text=#1,
    line width=0.35pt,
    minimum size=1.0em,
    inner sep=0pt,
    font=\bfseries\fontsize{6.5}{6.5}\selectfont
] {#2};
}%
}

\newcommand{\keyfinding}[1]{%
\begin{center}
\fcolorbox{blue!50!black}{blue!4}{%
\parbox{0.97\linewidth}{%
\small
\textbf{\color{blue!60!black}Key finding.} #1
}}
\end{center}
}

\newtcblisting{promptbox}[1]{%
  enhanced,
  breakable,
  listing only,
  listing engine=listings,
  listing options={%
    basicstyle=\ttfamily\scriptsize,
    breaklines=true,
    breakatwhitespace=false,
    columns=fullflexible,
    keepspaces=true,
    showstringspaces=false
  },
  colback=black!3,
  colframe=black!65,
  colbacktitle=black!65,
  coltitle=white,
  title={#1},
  fonttitle=\bfseries\small,
  arc=2mm,
  outer arc=2mm,
  boxrule=0.6pt,
  left=2.5mm,
  right=2.5mm,
  top=1.5mm,
  bottom=1.5mm,
  toptitle=1mm,
  bottomtitle=1mm,
  before skip=8pt,
  after skip=14pt
}

\newcommand{\promptheadingbreak}{\leavevmode\par\nopagebreak[4]}

\makeatletter
\newcommand{\daggerauthornote}[1]{%
  \authornotemark[2]%
  \g@addto@macro\@authornotes{\footnotetext[2]{#1}}%
}
\makeatother

\setcopyright{none}

\renewcommand\footnotetextcopyrightpermission[1]{}

\begin{document}

\title{\arena: Benchmarking LLMs on Real-World Predictions in Sports}

\author{Jonas Schr\"oder}
\daggerauthornote{Joint first authors.}
\affiliation{\institution{MCML \& LMU Munich} \city{Munich}\country{Germany}}
\email{jonas.schroeder@lmu.de}

\author{Jonas Schweisthal}
\authornotemark[2]
\affiliation{\institution{MCML \& LMU Munich}\city{Munich}\country{Germany}}
\email{jonas.schweisthal@lmu.de}

\author{Oliver M\"uller}
\affiliation{\institution{Paderborn University}\city{Paderborn}\country{Germany}}
\email{oliver.mueller@uni-paderborn.de}

\author{Markus Weinmann}
\affiliation{\institution{University of Cologne}\city{Cologne}\country{Germany}}
\email{weinmann@wiso.uni-koeln.de}

\author{Stefan Feuerriegel}
\affiliation{\institution{MCML \& LMU Munich}\city{Munich}\country{Germany}}
\email{feuerriegel@lmu.de}

\renewcommand{\shortauthors}{Schr\"oder and Schweisthal, et al.}

\begin{abstract}
Large language models (LLMs) increasingly support decisions about uncertain future events, yet evaluating their ability to forecast real-world outcomes remains difficult. In particular, existing benchmarks are typically static and retrospective, and therefore cannot test how information is synthesized by LLMs to predict future events under uncertainty. We introduce \textbf{\arena} (\arenalink), a prospective live benchmark that evaluates how well LLMs forecast real-world sports events before the outcomes are known. \arena provides \textbf{(1)} a prospective live benchmark protocol, \textbf{(2)} a public open-source platform, and \textbf{(3)} a factorial benchmark design together with tournament-related questions (e.g., which team will win). \arena automatically records timestamped, schema-validated forecasts of unresolved events, together with prompts, model versions, tool traces, and costs. The factorial design varies along four dimensions: \textbf{(1)} \emph{model version} (e.g., GPT-5.5, Claude Opus 4.8); \textbf{(2)} \emph{information access} (i.e., with or without web search); \textbf{(3)} \emph{prompting strategy} (i.e., scoreline or outcome probabilities), and \textbf{(4)} \emph{forecast horizon} (i.e., whether predictions are made at stage opening, 24 hours before kickoff, or 2 hours before kickoff). We demonstrate LLM-SoccerArena through a large-scale evaluation of the \emph{2026 FIFA World Cup}, in which seven LLMs generated forecasts for all 104 matches and 15 tournament-related questions.  We provide a detailed analysis of model performance across information access, prompting strategy, and forecast horizon. As a result, \arena provides new evidence about the forecasting performance of state-of-the-art LLMs. For example, LLMs with web access outperform those without, but only by a small margin (i.e., a 0.023 improvement in Brier score). Overall, \arena provides a flexible, open-source platform for prospective benchmarking of unresolved events. \arena will be continuously updated, and can be directly applied to future national and international tournaments and league competitions. 
\end{abstract}

\newcommand{\resourcebox}{%
  \par\noindent
  \begingroup
  \setlength{\fboxsep}{6pt}
  \setlength{\fboxrule}{0.9pt}
  \fcolorbox{draftblue}{blue!8}{%
    \parbox{\dimexpr\linewidth-2\fboxsep-2\fboxrule-3.2pt\relax}{%
      \raggedright\footnotesize

      {\color{draftblue}\textbf{Resources: The \soccerballicon\ \textbf{\arena} benchmark infrastructure}}

      \vspace{4pt}

      \trophyicon\hspace{0.5em}
      \textbf{Platform \& live leaderboard (continuously updated)}\\
      \url{https://www.llm-soccerarena.com/}

      \vspace{3pt}

      \clipboardicon\hspace{0.5em}
      \textbf{Dataset for the World Cup 2026 (CSV)}\\
      \url{https://github.com/jonas-srd/world_cup_LLM_rank/blob/main/data/worldcup2026-full-prediction-dataset-2026-07-21.csv}

      \vspace{3pt}

      \githubicon\hspace{0.5em}
      \textbf{Code \& reproducibility materials}\\
      \url{https://github.com/jonas-srd/world_cup_LLM_rank/tree/main}
    }%
  }%
  \endgroup
  \par\smallskip
}


\maketitle
\resourcebox

\section{Introduction}

Forecasting future events is important for decision-making in business, public policy, or personal life \cite{hyndmanForecastingPrinciplesPractice2018, PETROPOULOS2022705, hogarthForecastingPlanningEvaluation1981}. Recently, there has been a growing interest in whether large language models (LLMs) can provide such forecasts at scale \citep{autocast,halawi_etal_2024,forecastbench}. However, forecasting real-world events is challenging for various reasons: relevant information changes over time, outcomes remain unresolved, and models must reason under uncertainty based on incomplete evidence \citep{autocastpp,hsieh_etal_2024,gneiting_raftery_2007}. As such, forecasting with LLMs implicitly tests whether they can synthesize evolving information into well-founded probabilistic judgments about future events before the outcome is known.

Existing benchmarks for LLM-based forecasting have important limitations. Most are static and inherently \emph{retrospective}. This means they either evaluate questions for which outcomes are already known \cite{autocastpp, autocast} or which can be solved through reasoning from available information \cite{balunovic_etal_2025_matharena, dynabench, livebench}. In contrast, forecasting is fundamentally \emph{prospective}. It requires a model to predict an outcome that is still \emph{unresolved} using only the information available at the time of prediction. Hence, static benchmarks cannot reliably assess real-world forecasting ability because their questions and outcomes may already appear in model training data, causing performance to reflect memorization \citep{chen_etal_2025_contamination} or benchmark familiarity rather than real-world forecasting ability \cite{livebench, dynabench, dynaboard}. Further, asking a model to forecast from a historical date does not ensure that it ignores ex~ante knowledge acquired after that date \citep{exante}.

Dynamic benchmarks partly address this problem by continually adding, replacing, or updating tasks as models and training data change \citep{dynabench,dynaboard,livebench}. In particular, live forecasting benchmarks go further by collecting predictions before the corresponding outcomes are known \citep{forecastbench,prophetarena,tsarena}. For example, \emph{ForecastBench} \cite{forecastbench} covers broad questions drawn from prediction markets and continuously updated datasets; \emph{Prophet Arena} \cite{prophetarena} evaluates prediction-market events across politics, economics, entertainment, and science; and \emph{TS-Arena} \cite{tsarena} focuses on live numerical time series forecasts in the energy sector. However, continuous prospective evaluation remains difficult because existing benchmarks are constrained by the limited availability of standardized unresolved events. Moreover, the available event streams are often small, irregular, or heterogeneous, making repeated and controlled comparisons challenging.

In this paper, we introduce \textbf{\arena} (\arenalink), a general-purpose system for prospectively evaluating LLM forecasts of unresolved real-world sports events. A key strength of \arena is that it provides an automated, \emph{prospective} evaluation protocol to assess LLM forecasts before outcomes are known. \arena provides \circnum[tikzblue]{1} a prospective live benchmark protocol that registers unresolved events and records timestamped, schema-validated forecasts before their outcomes are known; \circnum[tikzblue]{2} a public \emph{open-source} platform that stores forecasts together with the corresponding prompts, model versions, reasoning traces, observed tool use, generated evidence (i.e., the LLM-generated rationale for the forecast), token use, and cost; and \circnum[tikzblue]{3} a factorial benchmark design to evaluate predictions of match results and of tournament-related questions (e.g., which team will win). The factorial benchmark design allows us to compare various settings that differ in \circnum[tikzgreen]{1}~\emph{model versions}  (e.g., GPT-5.5, Claude Opus 4.8), \circnum[tikzgreen]{2}~\emph{information access} (i.e., with or without web search); \circnum[tikzgreen]{3}~\emph{prompting strategy} (i.e., scoreline or outcome probabilities), and \circnum[tikzgreen]{4}~\emph{forecast horizon} (i.e., whether predictions are made at start of a tournament, 24 hours before a match, or 2 hours before a match). Further, \arena also supports a broad set of evaluation measures, including accuracy and Brier score.

\arena has several strengths as a general benchmark for LLM forecasting. \circnum[tikzred]{1}~It is fully \emph{prospective}: forecasts are recorded while outcomes remain unresolved, so performance reflects actual prediction (and prevents artifacts due to reconstruction, memorization, or access to known outcomes). \circnum[tikzred]{2}~It supports \emph{complex reasoning and information synthesis} in a highly \emph{standardized} real-world setting. Sports matches take place at scheduled times and produce objective, verifiable outcomes. The setting is further challenging because the relevant information changes continuously before kickoff and may include recent results, team rankings, historical matchups, injuries, expected lineups, venue information, weather, betting odds, and news. Models must therefore identify, retrieve, weigh, and combine heterogeneous evidence under uncertainty. More broadly, sports are often used in research as a proxy for studying complex management and decision-making problems. \circnum[tikzred]{3}~Soccer provides a \emph{challenging} forecasting problem: low score counts, draws, and unexpected results preserve substantial uncertainty even when teams differ in strength, which has motivated extensive statistical and machine learning research on match and tournament prediction \citep{maher_1982,dixon_coles_1997,groll_etal_2015,sportsprediction}. \circnum[tikzred]{4}~Sports provide a \emph{continuous, scalable} stream of comparable events. Our platform is interoperable with different league competitions (e.g., English Premier League, the German Bundesliga, and Spain’s La Liga), which together provide dozens of new forecasting events each week and thereby enable frequent benchmark updates. \circnum[tikzred]{5}~The benchmark can evaluate a broad range of currently relevant system capabilities, including probabilistic predictions (e.g., how well models can predict exact scores vs. outcome probabilities), calibration, and web search. Here, established models from sports analytics as well as bookmaker probabilities also provide external reference points for evaluating these forecasts \citep{leitner_etal_2010,ley_etal_2019,zeileis_etal_2018}. 

\arena is a \textbf{flexible} and fully \textbf{open-source} forecasting framework. \arena supports the complete real-time benchmark environment in an automated, end-to-end manner, from event registration and forecast scheduling to model execution, schema validation, archival, outcome retrieval, evaluation, and maintaining a live leaderboard. We make available not only the resulting predictions and evaluation data, but also the underlying platform (under a permissive MIT License; see our GitHub). Below, we demonstrate the system using the \emph{2026 FIFA World Cup}; however, the framework is designed to operate continuously across competitions, and we are currently extending it to cover major soccer leagues (e.g., the English \emph{Premier League}, the German \emph{Bundesliga}, where the tournaments launch in fall 2026) to offer a continuous, prospective benchmarking platform.

\begin{tcolorbox}[
enhanced,
breakable,
colback=draftlight,
colframe=draftblue,
boxrule=0.8pt,
arc=2pt,
left=6pt,
right=6pt,
top=5pt,
bottom=5pt,
title={Our contributions},
fonttitle=\bfseries,
coltitle=white,
colbacktitle=draftblue
]
\textbf{\circnum[tikzblue]{1} Prospective live benchmark protocol.}
We provide a prospective, real-time protocol that records and evaluates LLM forecasts for unresolved events before their outcomes are known.

\smallskip
\textbf{\circnum[tikzblue]{2} Public open-source platform.}
We provide a public platform under an open-source license, together with a live leaderboard and an auditable archive of past LLM forecasts and results.

\smallskip
\textbf{\circnum[tikzblue]{3} Factorial benchmark design.}
We compare LLM forecasts across different model versions, information access conditions, prompting strategies, and forecast horizons on the same events. Thereby, we derive recommendations for effective LLM use in real-world forecasts.

\smallskip
\textbf{\circnum[tikzblue]{4} Case study using the \emph{2026 FIFA World Cup}.}
We demonstrate \arena using on all 104 matches and 15 tournament questions of the \emph{2026 FIFA World Cup} and analyze forecast quality, reliability, and tool use. We also provide a qualitative analysis of the mentioned evidence that is provided by the LLM to justify a prediction.
\end{tcolorbox}

\section{Related Work}
Below, we position \arena relative to three relevant research streams (i.e., )\emph{LLM evaluation}, \emph{prospective LLM forecasting}, and \emph{soccer forecasting}) and explain how \arena is novel (see Table \ref{tab:related-work}).

\textbf{LLM evaluation.}
Standardized benchmarks and public evaluation platforms make comparisons across LLMs more transparent and reproducible by providing shared tasks, metrics, and evaluation procedures \citep{dynaboard, liang2023holisticevaluationlanguagemodels, chiang2024chatbotarenaopenplatform}. Existing benchmarks are often \emph{static} and therefore \emph{retrospective}. \cite{frauenNonparametricLLMEvaluation2026, autocast, autocastpp} Dynamic benchmarks address benchmark aging and training data contamination by collecting new examples or regularly updating tasks \citep{dynabench,livebench,balunovic_etal_2025_matharena,chen_etal_2025_contamination}. These approaches establish the importance of LLM evaluations that are \emph{public, repeatable, and continuously updated}. \textbf{\emph{However}}, dynamic benchmarks are often still retrospective: their  outputs can generally be evaluated against an existing answer or human judgment and therefore cannot test whether LLMs can forecast \emph{future} outcomes that are \emph{unknown}. 

\textbf{Prospective LLM forecasting.}
Prior work evaluates LLMs in \emph{prospective} forecasting tasks. One approach is represented by Autocast \citep{autocast} and Autocast++ \citep{autocastpp}, which reconstruct the information available at earlier dates using time-indexed news and retrieval to preserve the temporal order of information and reduce leakage from events occurring after the forecast date. \emph{\textbf{However,}} such retrospective reconstruction cannot fully prevent information leakage from post-forecast knowledge already encoded in the model and does not support benchmarking capabilities such as live agentic search and tool use under real-world conditions.

As second research stream provides live forecasting benchmarks, often in comparison with human forecasters \citep{halawi_etal_2024,pratt_etal_2024,hsieh_etal_2024}. Existing examples are scarce. ForecastBench \cite{forecastbench} and Prophet Arena  \citep{prophetarena} collect probabilistic forecasts for heterogeneous questions for which the outcomes are unresolved at forecast time, and TS-Arena \citep{tsarena} records live numerical forecasts for recurring energy data. As such, these benchmarks can establish the value of LLMs in \emph{prospective forecasting}. \textbf{\emph{However}}, they are often difficult to scale because unresolved events may be \emph{limited} or \emph{irregular}, and the tasks are heterogeneous and \emph{not} standardized (e.g., questions or events vary across time). In addition, benchmarks from this stream often do \emph{not} systematically evaluate prompting strategies or information access (e.g., web search) and provide limited qualitative insight into reasoning traces or evidence from LLM-generated justifications. \arena addresses this gap through a recurring stream of comparable soccer events evaluated under a standardized protocol.

\textbf{Soccer forecasting.}
Soccer forecasting traditionally uses curated domain information, such as historical results, team strength, rankings, player availability, and betting odds, to predict match scores, outcomes, or tournament progress \citep{maher_1982,dixon_coles_1997,groll_etal_2015,ley_etal_2019,zeileis_etal_2018}. Several works develop statistical and machine learning models specifically for this setting \cite{bunkerMachineLearningSoccer2024, rico-gonzalezMachineLearningApplication2022}. At the same time, bookmaker probabilities provide a strong external baseline that is often hard to beat \citep{leitner_etal_2010,sportsprediction}. \textbf{\emph{However}}, \arena has a different goal: we aim to evaluate the general-purpose forecasting abilities of LLMs rather than proposing a new specialized soccer model. We therefore deliberately do \emph{not} provide curated statistics or domain features beyond basic event information. This design keeps the benchmark simple and inexpensive to extend to new events and domains. It also allows us to directly test whether LLMs can synthesize relevant information and reason under uncertainty, and examine what additional information is retrieved through web search and how it is integrated into the forecasts.

\begin{table}[t]
\centering
\caption{Comparison of related research streams. \yes\ = supported, \partly\ = partially supported, \no\ = not supported.}
\label{tab:related-work}
\resizebox{\columnwidth}{!}{%
\begin{tabular}{p{4.6cm}ccccc}
\toprule
\textbf{Research stream}
& \textbf{Open-}
& \textbf{New unresolved}
& \textbf{Factorial}
& \textbf{Auditable forecast} & \textbf{Ex.} \\
& \textbf{source}
& \textbf{events}
& \textbf{benchmark}
& \textbf{records} \\
\midrule

Public LLM evaluation
{\scriptsize \citep{dynaboard, liang2023holisticevaluationlanguagemodels, chiang2024chatbotarenaopenplatform}
}
& \yes
& \no
& \no
& \partly \\

Dynamic LLM evaluation
{\scriptsize \citep{dynabench,livebench,balunovic_etal_2025_matharena}}
& \partly
& \no
& \no
& \partly \\

Live forecasting benchmarks
{\scriptsize \citep{forecastbench,prophetarena,tsarena}}
& \partly
& \yes
& \partly
& \partly \\
\midrule
\rowcolor{yellow!8}
\textbf{\arena}
& \yes
& \yes
& \yes
& \yes \\

\bottomrule
\end{tabular}%
}
\end{table}

\textbf{Research gap.} To the best of our knowledge, \arena is the first prospective benchmark for systematically evaluating LLMs on recurring real-world sports forecasts.

\begin{figure}[H]
    \centering
    \includegraphics[width=\columnwidth]{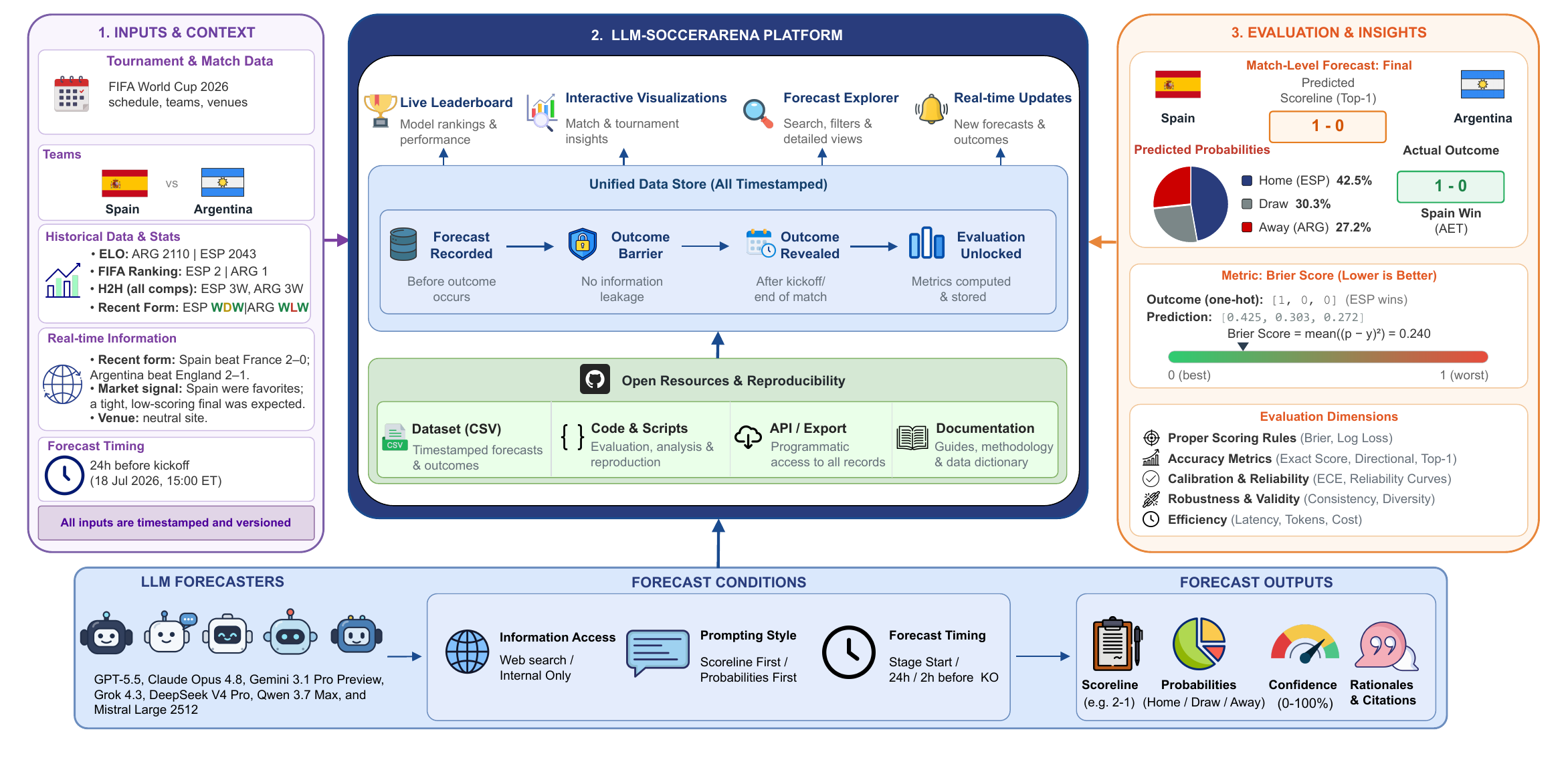}
    \caption{\arena architecture and public interface.}
    \label{fig:platform}
\end{figure}

\section{\arena}
\arena consists of three components (see Figure~\ref{fig:platform}): $\bullet$\,\textbf{\circnum[tikzblue]{1} Prospective live benchmark protocol:} to register unresolved events and record LLM forecasts before their outcomes are known (§\ref{sec:live-protocol}). $\bullet$\,\textbf{\circnum[tikzblue]{2} Public open source platform:} to provide forecasts, outcomes, data, code, and evaluation results using an auditable archive \cite{pineauImprovingReproducibilityMachine2020}. $\bullet$\,\textbf{\circnum[tikzblue]{3} Factorial benchmark design:} to compare selected model versions and other conditions (e.g., information access, prompting strategies, and forecast horizons) on the same event \cite{DesignAnalysisExperiments}. 

\subsection{Live Benchmark Protocol}
\label{sec:live-protocol}
The live benchmark protocol has two parts: (1)~The \emph{forecasting tasks} define what each LLM must predict. (2)~The \emph{prospective benchmark environment} defines how unresolved events and their forecasts are registered, collected, validated, stored, resolved, and evaluated.

\subsubsection{Forecasting Tasks}
\label{sec:forecasting-tasks}

We define two complementary forecasting tasks. (a)~\emph{Match forecasts} evaluate repeated short horizon predictions for individual matches. (b)~\emph{Tournament forecasts} evaluate structured predictions over longer horizons, such as group winners, semifinalists, or the champion.

\textbf{$\bullet$\,Match forecasts.}
For each match $i$, let $Y_i \in \mathcal{C} = \{\mathrm{H},\mathrm{D},\mathrm{A}\}$ denote the realized outcome after 90 min plus stoppage time (home win, draw, away win); extra time and penalty shootouts are excluded. The LLM outputs a probability vector
$\mathbf{p}_i = \bigl(p_{i,\mathrm{H}},\, p_{i,\mathrm{D}},\, p_{i,\mathrm{A}}\bigr) \in \Delta^{2},$
where $p_{i,c}$ is the probability the model assigns to the outcome $Y_i = c$ and where $\Delta^{2}$ denotes the probability simplex, i.e., $p_{i,c} \geq 0$ with $\sum_{c \in \mathcal{C}} p_{i,c} = 1$.

For a knockout match, the LLM additionally reports advancement probabilities, denoted by $a_{i,\mathrm{H}}$ and $a_{i,\mathrm{A}}$ with $a_{i,\mathrm{H}} + a_{i,\mathrm{A}} = 1$, where advancement accounts for extra time and penalty shootouts. We separate the 90 minute outcome from advancement because a knockout match can be drawn after 90 minutes even though exactly one team advances.

Each response also contains the expected number of goals for both teams, the most likely scoreline (e.g., 2–1), the model’s self-reported overall confidence on a continuous scale from 0 to 1, and a short generated rationale (which we refer to as \emph{evidence}). Predicted mean goals are not the football expected goals (``xG'') statistic. Confidence is the model's overall self assessment and is not treated as a probability that the selected outcome is correct. The generated evidence provides a rationale supporting the forecast and is therefore distinct from the model’s private reasoning traces. These additional information later allow us to evaluate how confident the model is and how the model reasoned over a forecast. 

\textbf{$\bullet$\,Tournament forecasts.}
Tournament forecasts are collected through natural-language questions over a predefined candidate set $\mathcal{T}_q$. Depending on the question, exactly one or exactly $k_q$ candidate teams are correct. For every candidate $j \in \mathcal{T}_q$, the LLM reports an associated probability together with its final prediction(s).

For one-team questions (e.g., ``Which team will win the FIFA World Cup?'') there is exactly one correct answer $Y_q \in \mathcal{T}_q$. The LLM reports a probability distribution
$\boldsymbol{\pi}_q$ over $\mathcal{T}_q$, where $\pi_{q,j}$ denotes the probability assigned to team $j$, together with one final selection.

Some questions instead require selecting exactly $k_q$ teams from the candidate set, such as predicting the four semifinalists. In this case, the LLM reports the marginal probability $\rho_{q,j}$ that team $j$ belongs to the correct set $S_q \subseteq \mathcal{T}_q$, subject to
$\rho_{q,j} \in [0,1], \;
\sum_{j\in\mathcal{T}_q}\rho_{q,j}=k_q,$
and outputs exactly $k_q$ final selections. The marginal probabilities sum to $k_q$ because exactly $k_q$ candidate teams are correct.

\subsubsection{LLM prompt}
For both tasks, we provide only basic information in prompt blocks shared across all benchmark conditions (details in Appendix \ref{appendix:prompts}). In particular, the prompt has two blocks: (1)~The  \emph{match block} contains the competition, edition, stage, UTC kickoff time, teams, known venue, and knockout status. (2)~The \emph{tournament block} contains the tournament structure, official fixtures, valid candidate teams, and the exact question definition. We deliberately do not provide curated historical results (e.g., form statistics, rankings, injuries, lineups, betting odds, or news). This design has two advantages. First, LLM forecasts can be collected without building and maintaining a separate structured data pipeline. Second, this allows us to evaluate the reasoning ability of web search, to assess which current information the LLM retrieves, synthesized, and included in the  generated evidence. Table~\ref{tab:forecast-fields} defines all recorded forecast fields and their interpretation. Appendix~\ref{app:prompts} provides the exact prompts, response schemas, candidate lists, and repair instructions.

\subsubsection{Prospective Benchmark Environment}
\label{sec:prospective-process}

The prospective benchmark environment separates forecast collection from resolving the outcome. This separation ensures that every LLM forecast is recorded under known conditions before the corresponding outcome becomes available. It proceeds in four steps:

\textbf{(1) Event registration.}
We register each match or tournament question while the corresponding outcome is unresolved. The registration contains a unique event identifier, the participating teams, the stage, the scheduled time, the available venue information, and the rule used to resolve the outcome. The same schema can register future tournaments and league fixtures, which creates a continually updated stream of unresolved events without changing the forecasting tasks.

\textbf{(2) Forecast collection.}
For each registered event, a cron-based scheduler automatically triggers the required forecast requests at prespecified forecast horizons. For match-level events, requests are targeted at $T-24$\,h and $T-2$\,h relative to the scheduled kickoff. Stage-opening forecasts are generated before the group stage begins or, for knockout stages, once all relevant pairings are known. The scheduler checks for due requests every 15 minutes and submits them through OpenRouter, which serves as a common API gateway to the registered model versions. Each request specifies the model version,
information access condition, prompting strategy, and forecast horizon. Because execution may deviate from the target time, we store both the scheduled and actual request timestamps and calculate the realized lead time as the interval between the actual request and kickoff. We additionally retain the exact prompt, raw response, parsed forecast, generated evidence, observed tool use, latency, token use, and cost.

\textbf{(3) Validation.}
A deterministic parser checks the response schema, required fields, the values of numbers, the ranges of probabilities, probability sums, score formats, and event-specific fields. Small deviations in probability vectors that must sum to one are normalized using a predefined rule and recorded. Other invalid responses receive at most one repair call. We retain the original response, every permitted transformation, API errors, timeouts, and unrepaired outputs. This procedure makes response reliability \emph{transparent} by allowing us to evaluate not only successful forecasts but also operational failures.

\textbf{(4) Outcome resolution.}
Official outcomes enter the benchmark only after the corresponding forecasts have been stored. This step acts as a barrier that prevents known outcomes from affecting the recorded forecasts. After resolution, we link each valid forecast to its outcome, calculate the evaluation measures (see Section \ref{sec:world-cup-setup}), and create a frozen analysis snapshot with integrity checks and hashes. The snapshot makes each benchmark release \emph{reproducible}, while later events and model versions can be evaluated through the same prospective live benchmark protocol.

Appendix~\ref{app:evaluation-details} provides the complete validation rules, archived fields, and snapshot checks. Appendix~\ref{app:prompts} provides the exact response schemas and repair prompts.
\vspace{-0.35em}
\subsection{Public Open-Source Platform}
\label{sec:public-platform}

The open-source platform (\arenalink) implements the live benchmark protocol and makes all records public. It provides {timestamped match and tournament forecasts, resolved outcomes, a filterable leaderboard, released data, and evaluation code}. The public website, released data, and analyses all use the same {auditable archive}. Every reported result can therefore be traced to the exact forecast, prompt, model version, benchmark conditions, validation record, observed tool use, evidence, latency, token use, and cost stored before the corresponding outcome was known. The website is accessed in a read-only mode, so that website interaction cannot modify stored forecasts or affect their evaluation. The released data retain both valid forecasts and operational failures, while the released code provides the corresponding evaluation procedures. Figure~\ref{fig:platform} summarizes the platform architecture.

\textbf{Leaderboard.} The website presents the LLM forecasts through four complementary views.
\emph{Match forecasts} display the predictions for individual matches, including probabilities, predicted scorelines, generated evidence, and benchmark conditions. \emph{Tournament questions} summarize longer-horizon forecasts, such as group winners, semifinalists, or the tournament champion. \emph{Bracket views} visualize the predicted tournament progression and allow comparisons with the realized tournament path. \emph{Performance summaries} aggregate evaluation metrics across models and benchmark conditions, providing rankings and interactive comparisons. Figure~\ref{fig:website} shows screenshots of the leaderboard.

\textbf{Future updates.} The platform is flexible and can be extended with \emph{new unresolved soccer events and new model versions}, including future tournaments and league fixtures, under the same live benchmark protocol. Together, these resources make \arena a \emph{public, continuously updated, auditable, and reproducible} benchmark. 

\subsection{Factorial Benchmark Design}
\label{sec:factorial-design}
The factorial benchmark design compares LLM forecasts under controlled combinations along four dimensions: \textbf{\circnum[tikzgreen]{1} model version}, \textbf{\circnum[tikzgreen]{2} information access}, \textbf{\circnum[tikzgreen]{3} prompting strategy}, and \textbf{\circnum[tikzgreen]{4} forecast horizon}. This design allows us to isolate how each factor contributes to forecast quality and to derive practical recommendations for using LLMs in real-world forecasting.
Let
\begin{equation}
\mathcal{Z}
=
\mathcal{M}\times\mathcal{A}\times\mathcal{P}\times\mathcal{H},
\qquad
z=(m,a,p,h)\in\mathcal{Z},
\label{eq:forecast-configurations}
\end{equation}
where $\mathcal{M}$, $\mathcal{A}$, $\mathcal{P}$, and $\mathcal{H}$
denote the sets of model versions, information access conditions,
prompting strategies, and forecast horizons, respectively.
For each registered match $i$, \arena creates forecasts for every
applicable configuration $z\in\mathcal{Z}$. All configurations receive
the same event information as input and produce the same structured
forecast outputs. Consequently, the benchmark dimensions modify only
the conditions under which a forecast is generated (i.e., the model
version, information access, prompting strategy, and forecast horizon),
while the forecasting task itself remains identical across all
configurations. This controlled design enables direct comparisons of
individual benchmark factors without confounding them with differences
in the underlying prediction task.
\textbf{(1) Model version.}
The model version identifies the exact model endpoint used to produce a forecast. For every forecast, we
record both the configured OpenRouter API identifier and the canonical frozen model version. Different model versions may vary in their numerical reasoning capabilities, reasoning strength, tool use. Here, we instantiate \arena with the following LLMs: GPT-5.5, Claude Opus 4.8, Gemini 3.1 Pro Preview, Grok 4.3, DeepSeek V4 Pro, Qwen 3.7 Max, and Mistral Large 2512. Comparing them on the same events and under identical conditions tests \emph{which model versions produce better forecasts}, rather than confounding model differences with event difficulty or deployment settings. Recording the exact endpoint also makes comparisons reproducible and allows future model versions to be evaluated under the same protocol. 

\textbf{(2) Information access.}
The information access condition tests whether knowledge encoded during LLM training is already sufficient for forecasting current events and whether access to recent external information improves the forecasts. (i)~In the \emph{closed-book} condition, web search is disabled and the LLM relies solely on its internal knowledge and the provided event information. (ii)~In the \emph{open-book} condition, web search is enabled and the LLM is instructed to retrieve current public information before forecasting. This comparison tests whether current information changes predictions and improves accuracy, as well as agentic web search can identify relevant information. For the subsequent analysis, we additionally record observed search use and the mentioned evidence to qualitatively study whether an open-book model actually retrieves relevant information and whether this information is reflected in its forecast.

\textbf{(3) Prompting strategy.}
The reasoning abilities of LLMs are known to deteriorate depending on whether qualitative or quantitative outputs are prompted \cite{atrejaWhatsPromptLargeScale2025}. Hence, by varying the prompting strategy, we can later test how LLMs can deal with events under uncertainty. (i)~In the \emph{scoreline} condition, the LLM is asked to report the most likely scoreline (and should then append the predicted mean goals and the outcome probabilities). (ii)~In the \emph{probabilistic forecast} condition, the LLM is asked to report the outcome probabilities and predicted mean goals (and should then combine these into the most likely scoreline). Both conditions use the same information, definitions, and response fields. Comparing the two conditions tests whether eliciting a concrete scoreline improves score prediction or whether eliciting probabilistic forecast improves probabilistic forecast quality.

\textbf{(4) Forecast horizon.}
The forecast horizon tests whether forecasts change as more recent event information becomes available and whether these changes improve prediction quality. Each match is forecast at three prespecified horizons. (i)~\emph{Stage opening} provides an early forecast before the corresponding tournament stage begins. (ii)~\emph{T--24h} provides the primary forecast approximately 24 hours before kickoff. (iii)~\emph{T--2h} provides a late forecast that may incorporate recent injuries, suspensions, expected lineups, and other team news. Comparing these horizons tests whether LLMs can effectively retrieve, synthesize, and reason over newly available information and thus become more accurate.

\textbf{Statistical analysis.}

For each comparison between two model versions or benchmark conditions (e.g., open versus closed book), we calculate one within-match metric difference and test whether its mean differs from zero using 10,000 two-sided sign-flip permutations \cite{frauenNonparametricLLMEvaluation2026,konietschkeBootstrappingPermutingPaired2014}. We apply Holm correction within prespecified families of related comparisons to correct for multiple comparisons \cite{holm1979}. Appendix~\ref{app:evaluation-details} provides further details about the statistical analysis.

\section{Case Study using the \emph{2026 FIFA World Cup}}
\subsection{Evaluation Setup}
\label{sec:world-cup-setup}
We demonstrate the abilities of \arena using \emph{2026 FIFA World Cup}. The tournament contains 104 matches among 48 teams and provides a sequence of match events with objective outcomes and rich information that can be used for web search. This setting allows us to evaluate all benchmark conditions repeatedly on the same forecast targets. Note that \arena is flexible and can be seamlessly adapted to other tournaments or sports leagues. 

\textbf{Model versions.} 
Our main evaluation includes seven state-of-the-art LLMs from different provider families: GPT-5.5, Claude Opus 4.8, Gemini 3.1 Pro Preview, Grok 4.3, DeepSeek V4 Pro, Qwen 3.7 Max, and Mistral Large 2512. The selection is intended to cover a diverse set of models from different providers. Table~\ref{tab:model-registry} reports a model registry with the exact API identifiers, provider names, and details on web access. 

\textbf{Match forecasts.} For every match, we apply the full factorial design with seven model versions, three forecast horizons, two information access conditions, and two prompting strategies. The main evaluation therefore contains $104 \times 7 \times 3 \times 2 \times 2 = 8{,}736$ match forecasts. Each LLM thus comes with 1,248 forecasts. This \emph{balanced and fully matched design} allows every model version and benchmark condition to be compared on the same matches without imputation. 

\textbf{Tournament forecasts.} Before the tournament began, the seven model versions answered 15 longer horizon questions: the 12 group winners, the four semifinalists, the world champion, and the team of the top scorer. Each question was evaluated under both information access conditions and both prompting strategies. The main evaluation therefore contains $15 \times 7 \times 2 \times 2 = 420$ tournament forecasts. We recorded these forecasts once at stage opening and froze them before the corresponding outcomes were known. We do not update them during the tournament because their purpose is to test whether LLMs can forecast the final tournament structure before it unfolds. 

\textbf{Performance Metrics.} We use different performance metrics to assess scoreline predictions, probability forecasts, and the accuracy of one-team questions. $\bullet$\,Our primary metric is the unscaled multiclass \textbf{Brier score}~[$\downarrow$]. For match $i$, let $\mathbf{p}_i$ denote the reported H/D/A probability vector and let $\mathbf{y}_i$ denote the one-hot encoding of the realized 90-minute outcome. We additionally report the \textbf{log loss}~[$\downarrow$]. The two metrics are defined as
$\mathrm{BS}_i = 
   \sum_{c\in\mathcal{C}}
     \left(p_{i,c}-y_{i,c}\right)^2,
\;
\mathrm{LL}_i = 
   -\log p_{i,Y_i}.$

Lower values indicate better forecasts. A perfect forecast has a Brier score of zero, the worst possible forecast has a score of two, and a uniform forecast has a score of $2/3$. We use the Brier score because it evaluates the \emph{complete probability distribution}, rather than only the outcome with the highest probability \citep{brier_1950,gneiting_raftery_2007}. Log loss penalizes confident incorrect forecasts more strongly. $\bullet$\,The \textbf{modal H/D/A accuracy}~[$\uparrow$] measures whether the realized outcome has the highest reported probability; if $k$ outcomes share the maximum probability, each receives credit $1/k$. $\bullet$\,The \textbf{exact-score accuracy}~[$\uparrow$] measures whether the reported scoreline equals the realized 90-minute score. $\bullet$\,The \textbf{Scoring System}~[$\uparrow$] provides a graded scoreline measure: five points are awarded for the exact score, two for the correct goal difference, one for the correct H/D/A tendency, and zero otherwise.\footnote{The scoring system is inspired by \url{https://www.kicktipp.de/}.}

For one-team questions, such as the world champion, we report Brier score, log loss, and accuracy. For the semifinalist question, we also report whether all four teams are correct, how many semifinalists are correct, the mean squared error between each team's reported probability of reaching the semifinals and the observed outcome, and whether these probabilities sum to four. Table~\ref{tab:evaluation-metrics} defines all measures.

We organize the following results around five evaluation dimensions: \textbf{(1) model performance}, \textbf{(2) information access}, \textbf{(3) prompting strategy}, \textbf{(4) calibration}, and \textbf{(5) tournament forecasts}. Unless stated otherwise, match results use the balanced seven-model panel and compare forecasts on the same matches under matched benchmark conditions.

\begin{table*}[t]
  \centering
\caption{Complete-panel T--24h probabilistic forecast performance. Values are means with 95\% match-bootstrap intervals.}
  \label{tab:leaderboard}

  \scriptsize
  \setlength{\tabcolsep}{1.2pt}
  \renewcommand{\arraystretch}{1.05}

  \resizebox{\textwidth}{!}{%
  \begin{tabular}{@{}p{3.35cm}rrrrrrr@{}}
  \toprule
  \textbf{Model} &
  \textbf{Brier [95\% CI]}$\downarrow$ &
  \textbf{Log loss [95\% CI]}$\downarrow$ &
  \textbf{Modal H/D/A accuracy, \% [95\% CI]}$\uparrow$ &
  \textbf{Exact score, \% [95\% CI]}$\uparrow$ &
  \textbf{Scoring System [95\% CI]}$\uparrow$ &
  \textbf{n} \\
  \midrule

  \googleicon Gemini 3.1 Pro Preview &
  0.506 [0.433, 0.587] &
  0.853 [0.757, 0.963] &
  63.7 [54.0, 72.6] &
  15.4 [9.8, 22.8] &
  1.36 [1.10, 1.68] &
  104 \\

  \openaiicon GPT-5.5 &
  0.517 [0.454, 0.584] &
  0.873 [0.789, 0.963] &
  62.0 [52.4, 70.9] &
  15.4 [10.0, 22.5] &
  1.39 [1.12, 1.70] &
  104 \\

  \deepseekicon DeepSeek V4 Pro &
  0.520 [0.452, 0.596] &
  0.877 [0.781, 0.980] &
  63.7 [54.2, 72.4] &
  13.9 [9.0, 20.6] &
  1.27 [1.02, 1.56] &
  104 \\

  \xaiicon Grok 4.3 &
  0.524 [0.455, 0.600] &
  0.876 [0.781, 0.979] &
  63.5 [53.6, 72.4] &
  11.5 [7.0, 18.1] &
  1.19 [0.96, 1.48] &
  104 \\

  \alibabaicon Qwen 3.7 Max &
  0.527 [0.463, 0.597] &
  0.887 [0.803, 0.978] &
  61.1 [51.1, 70.1] &
  13.5 [8.6, 20.3] &
  1.24 [0.99, 1.54] &
  104 \\

  \anthropicicon Claude Opus 4.8 &
  0.528 [0.465, 0.597] &
  0.888 [0.803, 0.982] &
  63.0 [53.2, 71.8] &
  14.9 [9.5, 22.3] &
  1.25 [0.98, 1.57] &
  104 \\

  \mistralicon Mistral Large 2512 &
  0.546 [0.493, 0.604] &
  0.913 [0.839, 0.989] &
  59.6 [50.2, 68.1] &
  12.5 [7.6, 19.2] &
  1.17 [0.93, 1.47] &
  104 \\

  \bottomrule
  \end{tabular}}
\end{table*}

\subsection{Model Performance}
\textbf{Forecast quality.}
Overall, the seven LLMs exhibit comparable forecasting performance, with small differences depending on the evaluation metric. Gemini achieves the best probabilistic forecasts, obtaining the lowest Brier score and log loss, whereas other models perform best on complementary metrics such as \emph{modal H/D/A} accuracy, exact score accuracy, or Scoring System points. Across models, Brier scores range only from $0.506$ to $0.546$. Consistent with these small differences, none of the 21 paired model comparisons remains statistically significant after Holm correction (smallest adjusted $p=0.055$ for GPT versus Mistral). The absence of statistically significant differences is unsurprising given the sample of 104 matches and is consistent with findings from other prospective LLM forecasting benchmarks. 
Table~\ref{tab:leaderboard} reports the complete results, while Figure~\ref{fig:overall} in Appendix~\ref{app:performance-details} shows cumulative and stage-specific performance over the tournament.

The comparison of open-book, probabilistic T--2h forecasts with de-vigged closing bookmaker odds
shows competitive absolute performance: Gemini 3.1 Pro Preview achieves a mean Brier score of $0.497$, compared with $0.498$ for the market consensus. Figures~\ref{fig:closing-odds-absolute} and~\ref{fig:closing-odds-paired} in the appendix report the absolute and paired comparisons.

\textbf{Forecast diversity.}
We now study whether forecasts are similar or correlated. Within the same match and benchmark conditions, the seven LLMs assign highly correlated H/D/A probabilities, with a mean pairwise correlation of $0.943$. The corresponding mean Jensen--Shannon divergence is only $0.0044$. The models therefore provide \emph{largely overlapping forecast signals}, rather than independent forecasts of the same event.

We thus test whether the across-model diversity is useful through an equal-weight ensemble. The ensemble improves on the average member by only $0.0047$ Brier, which directly reflects the low disagreement among the forecasts, and none of the seven ensemble-–model comparisons remains significant after Holm correction. Aggregation therefore provides \emph{only a small gain because the underlying forecasts are already highly similar}. Figure~\ref{fig:forecast-diversity} in Appendix~\ref{app:performance-details} reports the pairwise similarity matrices and ensemble contrasts. The same appendix provides the complete pairwise tests and results by tournament stage.

\keyfinding{The leaderboard shows no clear winner, and the seven models produce highly similar forecasts.}

\subsection{Information Access}
\begin{figure}[t]
  \centering

  \begin{minipage}[c]{0.48\columnwidth}
    \centering
    \begin{tabular}{@{}c@{\hspace{0.15em}}c@{}}
      \textbf{(a)} &
      \raisebox{-0.5\height}{%
        \includegraphics[
          width=0.86\linewidth,
          trim=0 2mm 0 2mm,
          clip
        ]{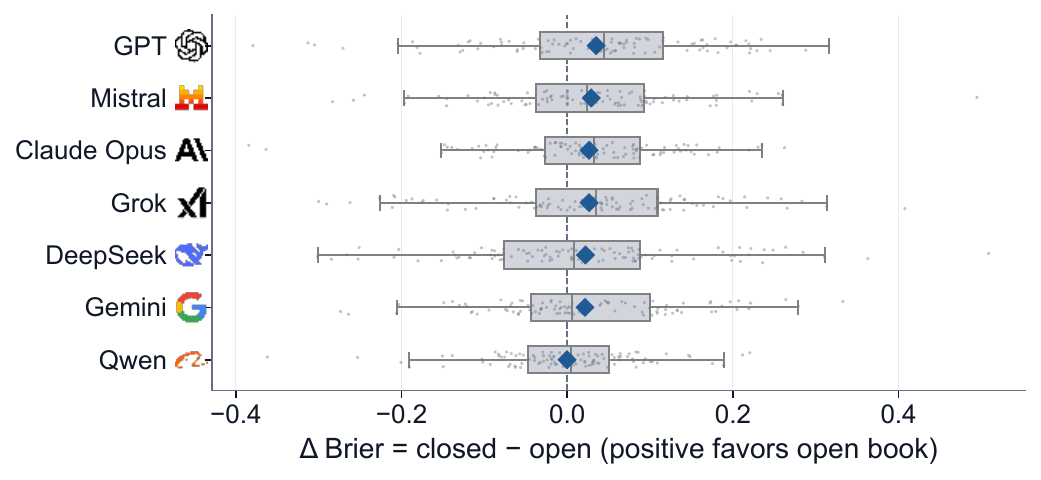}%
      }
    \end{tabular}

    \vspace{0.25em}

    \begin{tabular}{@{}c@{\hspace{0.15em}}c@{}}
      \textbf{(b)} &
      \raisebox{-0.5\height}{%
        \includegraphics[
          width=0.86\linewidth,
          trim=0 2mm 0 2mm,
          clip
        ]{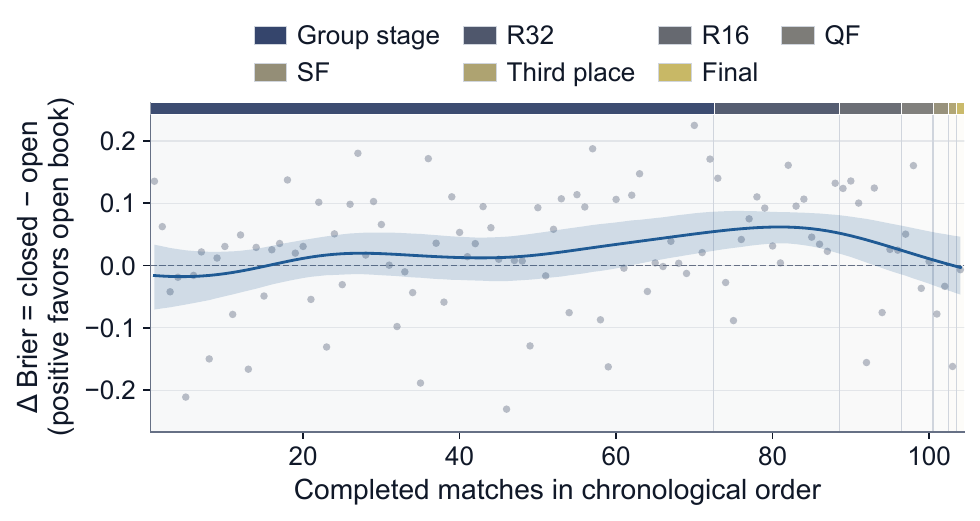}%
      }
    \end{tabular}
  \end{minipage}
  \hfill
  \begin{minipage}[c]{0.48\columnwidth}
    \centering
    \begin{tabular}{@{}c@{\hspace{0.15em}}c@{}}
      \textbf{(c)} &
      \raisebox{-0.5\height}{%
        \includegraphics[
          width=0.86\linewidth,
          trim=0 2mm 0 2mm,
          clip
        ]{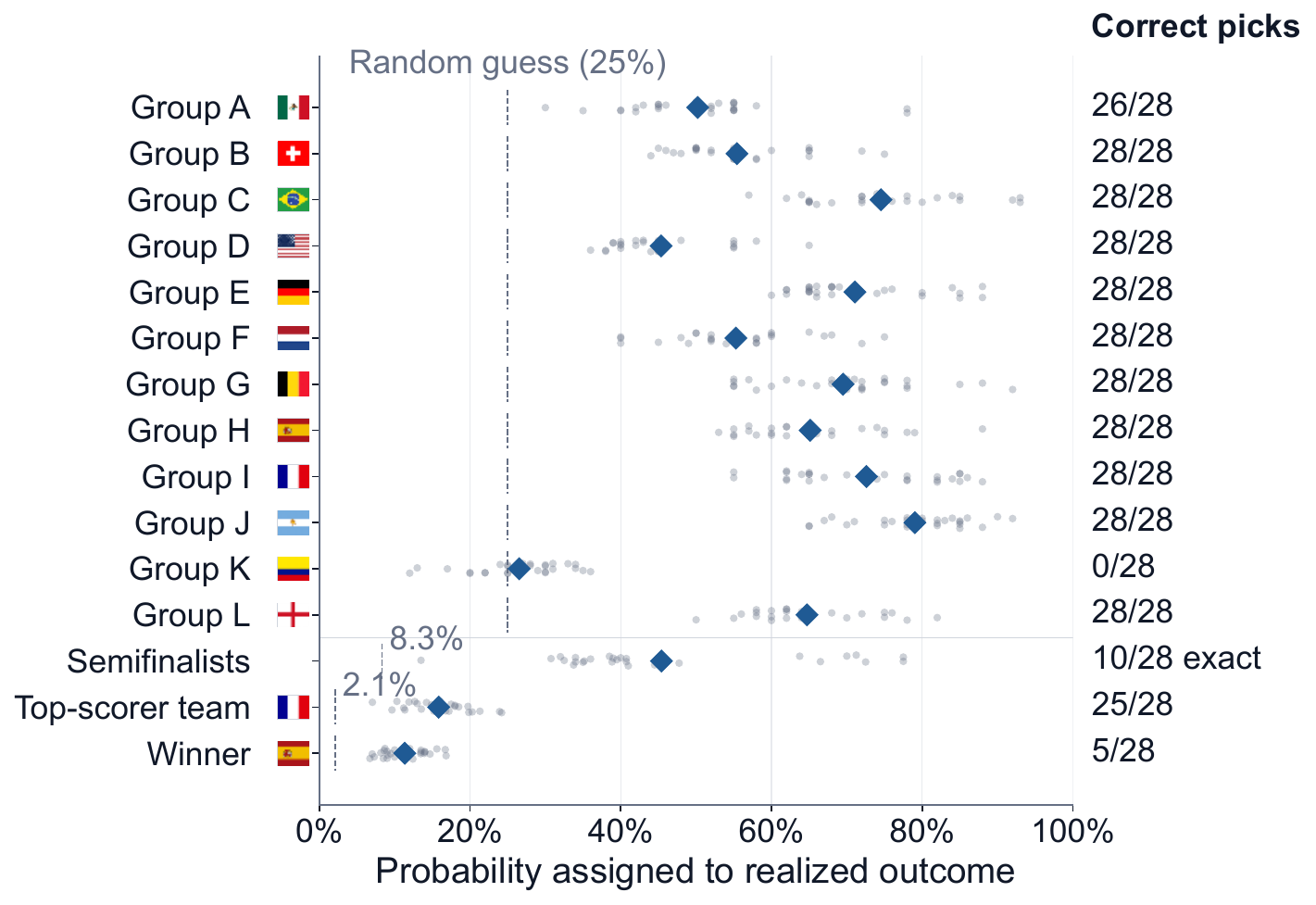}%
      }
    \end{tabular}

    \vspace{0.25em}

    \begin{tabular}{@{}c@{\hspace{0.15em}}c@{}}
      \textbf{(d)} &
      \raisebox{-0.5\height}{%
        \includegraphics[
          width=0.86\linewidth,
          trim=0 2mm 0 2mm,
          clip
        ]{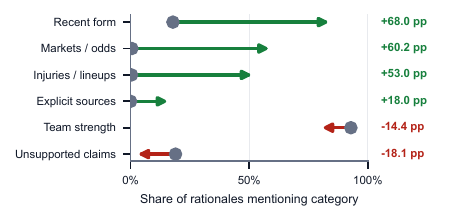}%
      }
    \end{tabular}
  \end{minipage}

  \vspace{-0.4em}
\caption{\textbf{(a)} Model-specific and \textbf{(b)} match-level effects of open-book access. \textbf{(c)} Tournament-question forecasts. \textbf{(d)} Evidence-category shares open-book vs. closed-book.}
  \label{fig:combined-results}
\end{figure}

\textbf{Forecast quality.}
Open-book access provides the clearest controlled improvement in forecast quality. At T--24h, the mean Brier score decreases from $0.535$ in the closed-book condition to $0.512$ in the open-book condition. The paired closed-minus-open difference is $0.0228$, with a $95\%$ confidence interval of $[0.0044, 0.0403]$ and a Holm-adjusted $p$-value of $0.045$. This corresponds to a $4.3\%$ reduction from the closed-book mean. The information access effect is significant after Holm correction, whereas the 21 pairwise model comparisons indicate similar performance across the seven models. Information access therefore represents a more important source of variation in \emph{forecast quality} than the observed differences among model versions. Figure~\ref{fig:combined-results}\textbf{(a)} shows the model-specific effects, while panel~\textbf{(b)} shows how the effect develops over the tournament (i.e., hinting toward a slight upward trend).

\textbf{Forecast horizon.}
More recent open-book forecasts (i.e., T--2h instead of T--24h) improve only slightly. The mean Brier score decreases by $0.0054$ from stage opening to T--24h and by $0.0021$ from T--24h to T--2h. Closed-book forecasts change little across the same horizons. Thus, web access improves forecast quality, but \emph{forecasting closer to kickoff does not by itself improve performance}.

\textbf{Search use.}
Interestingly, not all LLMs with enabling web search eventually use it. We observe search functionality only in $84.5\%$ of open-book forecasts, while model-specific rates range from $48.4\%$ to $100\%$. Open-book access adds on average 22,306 input tokens, 885 output tokens, 3.92 seconds of latency, and USD~0.110 per forecast. Figure~\ref{fig:access-details} in Appendix~\ref{app:access-details} reports the pooled access effect, forecast horizon comparisons, forecast changes across snapshots, and operational costs. 

\textbf{Mentioned evidence.}
We qualitatively analyze the LLM-generated evidence from 1,456 forecasts using the T--24h forecast horizon and the probabilistic forecast prompting strategy. We analyze the evidence using the following prespecified categories: \emph{markets or odds}, \emph{recent form}, \emph{injuries or lineups}, \emph{rankings or team strength}, \emph{tactics}, \emph{venue or travel}, \emph{tournament context}, \emph{explicit sources}, and \emph{generic unsupported claims}. For this, we use a released keyword lexicon together with a frozen GLM~5.2 annotator to label the rationales. Following recommendations in \cite{feuerriegelUsingNaturalLanguage2025}, we cross-checked the labels using a blinded human audit of 196 rationales. Figure~\ref{fig:rationale-evidence-complete} in Appendix~\ref{app:rationale-analysis} reports detailed results across different evidence categories and model-specific patterns.

Compared with closed-book rationales, open-book rationales mention \emph{recent form} 68.0 percentage points more often, \emph{markets or odds} 60.2 percentage points more often, and \emph{injuries or lineups} 53.0 percentage points more often. Conversely, \emph{generic unsupported claims}---predictive assertions or football clich\'es without a concrete factual or mechanistic basis in the rationale---occur 18.1 percentage points less often. All seven LLMs show the same directional patterns. 
Details are in Figure~\ref{fig:combined-results}\textbf{(d)}. Appendix~\ref{app:access-details} reports robustness and observed-search analyses and decomposes Brier-score variation across models and conditions. Appendix~\ref{app:rationale-analysis} details the annotation
protocol and human audit.
\keyfinding{Web access improves forecasts, but forecasting closer to kickoff adds little and models use search very differently.}

\subsection{Prompting Strategy}
We find no evidence that prompt order changes average forecast quality. For each match, we compare Brier scores from score-first and probabilistic forecast under otherwise identical conditions. The mean difference, calculated as score first minus probabilistic , is $-0.0008$, with a 95\% confidence interval of $[-0.0044, 0.0029]$ and a Holm-adjusted $p$-value of $0.693$. Positive values would favor probabilistic forecast.

\textbf{Forecast structure.}
Compared with score-first prompting, probabilistic forecast produces 3.98 percentage points more predicted draw scorelines and
4.05 percentage points more $1$--$1$ scorelines, while the mean probability assigned to a draw is 0.21 percentage points lower.
Six of the seven model versions show an increase in predicted draw scorelines.

Details are in Appendix~\ref{app:prompting-details} and Table~\ref{tab:output-validity} in Appendix~\ref{app:evaluation-details}.

\keyfinding{Prompt order does not change accuracy, but probabilistic forecast produces more draw scorelines and more disagreement with outcome probabilities.}

\subsection{Analysis for Calibration and Confidence}

We evaluate two reported outputs. First, we test the calibration, that is, whether the H/D/A probabilities match observed frequencies (i.e., whether outcomes assigned $60\%$ occur about $60\%$ of the time). Second, we test whether higher self-reported confidence (i.e., the model's overall self-assessment) is linked to more accurate forecasts. 
\textbf{(1) Calibration.}
For each T--24h forecast horizon, information access condition, and prompting strategy, we average the probability vectors of the seven model versions. The calibration curves broadly follow the diagonal, but several probability ranges deviate.

\textbf{(2)~Confidence.}
A pooled correlation would conflate model-specific confidence scales and treat repeated forecasts for the same match as independent.
We therefore rank confidence within each model-version, information-access, and prompting-strategy cell and compare relative confidence groups. From the lowest to the highest group, the mean Brier score decreases from $0.620$ to $0.423$, while modal H/D/A accuracy increases from $49.8\%$ to $73.1\%$. Higher relative confidence is thus
associated with better forecasts, but is not a calibrated probability of correctness. Figure~\ref{fig:calibration} and Appendix~\ref{app:calibration-details} provide the detailed results.
\keyfinding{Probability calibration shows no consistent bias, while higher self reported confidence identifies more accurate forecasts.}

\section{Discussion}
\label{sec:discussion}

\textbf{Strengths.}
\arena offers several advantages as a benchmark setting for real-world LLM forecasting. \textbf{(1)}~It is fully \emph{prospective} yet \emph{scalable}: forecasts are recorded \emph{before} outcomes are known, while recurring tournaments and league competitions provide a continuous stream of events with standardized outcomes for evaluating new LLMs over time. \textbf{(2)}~Sports provide a well-established empirical setting for studying broader management and organizational phenomena, including decision-making, competition, collaboration, leadership, and performance under uncertainty \cite{wolfe2005, day2012}. Hence, the findings can inform recommendations for effective LLM prompting in practice. The standardized format of sports outcomes further enables direct comparisons between LLMs, but also against specialized statistical and machine learning models, human forecasts, prediction markets, and bookmaker-implied probabilities on the same outcomes. \textbf{(3)}~Forecasting soccer results presents a challenging test of information retrieval and synthesis; web-enabled agents must distinguish relevant evidence from noise and conflicting reports, account for uneven coverage across countries and languages, and integrate signals such as form, injuries, lineups, weather, and market information. The collected rationales and observable tool traces provide qualitative insights into the evidence that LLMs use to support their forecasts. \textbf{(4)} The complete benchmark platform is open source.

\textbf{Practical implications for LLM prompting and deployment.}
Our results suggest three concrete recommendations for managers deploying LLM forecasts. \textbf{(1)}~\emph{Provide web access when decisions depend on current information rather than relying solely on a larger or newer model.} Our results show that open-book forecasts outperform closed-book forecasts, whereas model differences are small. However, practitioners should verify actual search use and record the retrieved evidence, because access to search does not ensure that the model uses it. \textbf{(2)}~\emph{Prompt design can benefit from probabilistic approaches.}  Score-first forecasts may be easier to communicate, but in our results, score-first prompting did not improve predictive accuracy (and only increased consistency across the reported outputs). For decision support \cite{powerDecisionSupportSystems2002,shimPresentFutureDecision2002}, probability distributions may be more useful because they make uncertainty and alternative outcomes explicit. In practice, this could imply that managers prompt LLMs to construct several plausible scenarios, assess their relative likelihoods, and combine them into a coherent probabilistic forecast rather than requesting only a single prediction. \textbf{(3)}~\emph{Self-reported confidence should be used carefully.} Although practitioners may be tempted to ask an LLM how confident it is, similar to how managers consult human experts, the reported confidence is not necessarily well calibrated. In some cases, it may help prioritize relatively easier or harder tasks, but it should not be interpreted as a signal whether the forecast is correct.

\textbf{Limitations.} As with any benchmark, \arena has limitations. First, our case study covers only one tournament, even though the \emph{FIFA World Cup} is a high-profile, globally followed competition with a large and diverse set of matches and large attention with an audience of more than one billion viewers. Our platform itself is flexible, and we are currently extending it to regularly cover additional competitions, including the English \emph{Premier League} and the German \emph{Bundesliga}. Second, analyzing the generated justifications and mentioned evidence by LLMs provides useful qualitative insights, but these outputs do not reveal private model reasoning or establish which information causally influenced a forecast. Third, we deliberately use simple and standardized prompts to ensure comparability and to test how LLMs benefit from web searches. This provides many opportunities for future research to examine alternative prompting strategies, more advanced agent architectures, retrieval procedures, and additional forms of tool use. Fourth, similar to humans, LLM forecasts are not perfect; however, this also reflects the inherent difficulty of soccer forecasting, where substantial uncertainty, draws, and unexpected outcomes persist even for specialized statistical models and betting markets. As such, the irreducible uncertainty is precisely what makes soccer a challenging setting for evaluating real-world forecasting capabilities.

\textbf{Research opportunities.}
\arena creates several concrete opportunities for future research. \textbf{(1)}~The design as a continuous benchmark enables longitudinal studies of whether new model generations improve in forecast accuracy, calibration, search behavior, and cost over time. \textbf{(2)}~The standardized setting allows researchers to test how prompting strategies and agent designs affect numerical reasoning. For example, researchers could test whether probabilistic forecasts could be improved through more structured prompting strategies that require intermediate calculations or consistency checks. \textbf{(3)} The recorded evidence and tool traces enable systematic analysis of which search and reasoning behaviors are associated with better forecasts. For example, models sometimes retrieve archival or otherwise weakly relevant sources (e.g., \texttt{arxiv.org}), which makes it possible to study how source selection, evidence quality, and prompting interventions affect forecasting performance and how these behaviors change across model generations.

\textbf{Conclusion.}
We propose \arena, an open-source, prospective benchmark and live leaderboard for evaluating LLM forecasts of real-world events where the outcomes are unknown. We demonstrate the benchmark through a case study of the \emph{2026 FIFA World Cup}, while the design of \arena is flexible and can easily be extended to other tournaments, leagues, and model versions. To the best of our knowledge, \arena is the first benchmark to provide such a continuously operating and standardized evaluation of LLM forecasting in sports.

\clearpage

\appendix

\section*{Disclosure of LLM use}

OpenAI Codex and Anthropic Claude were used as assistive tools for software development, analysis
implementation, visualization, and manuscript drafting and editing; the authors conceived the study,
directed and verified the work, and take responsibility for the final content.

\section{Future Deployment}

\begin{center}
\fcolorbox{black!25}{black!4}{%
\begin{minipage}{\dimexpr\columnwidth-2\fboxsep-2\fboxrule\relax}
\textbf{Long-term operation, maintenance, and extensions.}
\arena is designed as a continuously operating benchmark. Its automated end-to-end pipeline supports event registration, forecast collection, validation, outcome resolution, evaluation, and leaderboard updates with limited manual intervention. We plan to extend the platform to the 2026/27 seasons of \emph{La Liga}, beginning on 15--16 August 2026; the \emph{Premier League}, beginning on 21 August 2026; and the \emph{Bundesliga}, beginning on 28 August 2026. Long-term maintenance includes regularly registering new model versions and forecasting configurations. Continuous operation enables evaluation at scale and create a longitudinal record of changes in LLM forecasting, tool use, and probabilistic reasoning.
\end{minipage}%
}
\end{center}

\section{Evaluation and Reproducibility Details}
\label{app:evaluation-details}

This appendix provides the details required to reproduce the World Cup 2026 evaluation. It defines the frozen analysis snapshot, model registry, recorded fields, evaluation measures, statistical procedures, and supplementary analyses. The main results use the balanced seven-model panel without imputation. Claude Fable is retained only as additional archive coverage and is excluded from all main analyses. Reporting follows \cite{feuerriegel2026reporting}.

\subsection{Frozen Analysis Snapshot}

We create the analysis snapshot with SQLite's backup mechanism inside a read transaction. The pipeline runs \texttt{PRAGMA integrity\_check} and records the UTC freeze time, database SHA-256, schema version, table counts, source path, and repository commit. Final-paper mode requires 104 reconciled fixtures and one official 90-minute outcome for every completed match.

We derive typed Parquet tables with explicit schemas and UTC timestamps. Each table records the source-database hash, configuration hash, code commit, row count, key-uniqueness result, and file hash. We read the public website CSV once for reconciliation. Any discrepancy is logged explicitly; the pipeline never resolves conflicts by silently preferring one source.

\subsection{Frozen Model Registry}

Table~\ref{tab:model-registry} resolves every short model name used in the paper. All calls were routed through OpenRouter to the listed provider endpoint. The frozen registry did not store a verified public-weight release for any exact deployed endpoint, so the table reports ``unverified'' rather than inferring availability from a related model family. Web search was assigned in the open-book condition and disabled in the closed-book condition. Observed search use is analyzed separately.

\begin{table*}[t]
  \centering
  \caption{Model registry. The seven complete model versions form the balanced main analysis panel. Note: the online archive also includes Claude Fable 5, but it has partial coverage because data collection was temporarily paused due to U.S. sanctions.}
  \label{tab:model-registry}
  \scriptsize
  \setlength{\tabcolsep}{3pt}
  \renewcommand{\arraystretch}{1.12}
  \begin{tabularx}{\textwidth}{@{}p{2.35cm}p{1.55cm}p{3.25cm}p{3.75cm}p{1.25cm}p{1.20cm}X@{}}
    \toprule
    \textbf{Model} & \textbf{Provider} & \textbf{Configured API identifier} &
    \textbf{Canonical frozen version} & \textbf{Public weights} & \textbf{Web access} &
    \textbf{Analysis role} \\
    \midrule
    GPT-5.5 & OpenAI & \path{openai/gpt-5.5} & \path{openai/gpt-5.5-20260423} & Unverified & Assigned & Main panel \\
    Claude Opus 4.8 & Anthropic & \path{anthropic/claude-opus-4.8} & \path{anthropic/claude-4.8-opus-20260528} & Unverified & Assigned & Main panel \\
    Gemini 3.1 Pro Preview & Google & \path{google/gemini-3.1-pro-preview} & \path{google/gemini-3.1-pro-preview-20260219} & Unverified & Assigned & Main panel \\
    Grok 4.3 & xAI & \path{x-ai/grok-4.3} & \path{x-ai/grok-4.3-20260430} & Unverified & Assigned & Main panel \\
    DeepSeek V4 Pro & DeepSeek & \path{deepseek/deepseek-v4-pro} & \path{deepseek/deepseek-v4-pro-20260423} & Unverified & Assigned & Main panel \\
    Qwen 3.7 Max & Qwen & \path{qwen/qwen3.7-max} & \path{qwen/qwen3.7-max-20260520} & Unverified & Assigned & Main panel \\
    Mistral Large 2512 & Mistral AI & \path{mistralai/mistral-large-2512} & \path{mistralai/mistral-large-2512} & Unverified & Assigned & Main panel \\
    \bottomrule
  \end{tabularx}
\end{table*}

\subsection{Recorded Forecast Fields}

Table~\ref{tab:forecast-fields} defines the content and provenance stored for every match forecast. All benchmark conditions use the same response schema. Information access and prompting strategy change how the forecast is produced, not what the model must report.

\begin{table*}[t]
  \caption{Forecast content and provenance recorded for each match-level attempt.}
  \label{tab:forecast-fields}
  \centering
  \small
  \begin{tabularx}{\textwidth}{>{\raggedright\arraybackslash}p{0.20\textwidth}
    >{\raggedright\arraybackslash}p{0.31\textwidth}X}
    \toprule
    \textbf{Object} & \textbf{Stored fields} & \textbf{Interpretation and use} \\
    \midrule
    90-minute probabilities &
    $p_{i,\mathrm{H}}$, $p_{i,\mathrm{D}}$, and $p_{i,\mathrm{A}}$ &
    Probabilities of a home win, draw, and away win after regulation time plus stoppage time. Values must lie in $[0,1]$ and sum to one. This is the primary forecast target. \\

    Goal and score forecast &
    Predicted mean home and away goals; most likely 90-minute score &
    Predicted mean goal counts and one most likely scoreline. These fields support exact-score, goal-difference, tendency, error, and scoreline-probability agreement analyses. \\

    Advancement forecast &
    Home and away advancement probabilities &
    For knockout matches, the probability that each team advances after extra time and penalties if required. For group-stage matches, these fields are null. \\

    Self assessment and rationale &
    Confidence in $[0,1]$; short generated rationale &
    Confidence is the model's overall self assessment, not a probability that its selected outcome is correct. The rationale supports analysis of mentioned evidence and is not treated as private model reasoning. \\

    Benchmark identity &
    Match, model version, provider, forecast horizon, information access, prompting strategy, sample ID &
    These fields form the stable forecast key and prevent accidental duplication or mixing of benchmark conditions. \\

    Execution provenance &
    Scheduled and actual timestamps, minutes to kickoff, prompt and template hashes, raw response, response ID, tool trace, latency, tokens, and cost &
    These fields make timing, observed search use, failures, and resource use auditable without reconstructing them from the public website. \\

    Validation provenance &
    Parsed fields, validation status and errors, original and final probability sums, normalization and repair flags, scoring eligibility &
    Raw responses are never overwritten. Validated values and every permitted transformation remain linked to the original response. \\
    \bottomrule
  \end{tabularx}
\end{table*}

\subsection{Evaluation Measures}

Table~\ref{tab:evaluation-metrics} defines the complete evaluation set. We keep the measures separate because probability quality, categorical accuracy, score prediction, confidence, output validity, and operational behavior capture different properties of an LLM forecast.

\begin{table*}[t]
  \caption{Evaluation measures. Lower is better for losses and errors; higher is better for accuracy, points, validity, and observed-search rates. Calibration, confidence, agreement, diversity, and rationale analyses are diagnostic.}
  \label{tab:evaluation-metrics}
  \centering
  \scriptsize
  \begin{tabularx}{\textwidth}{>{\raggedright\arraybackslash}p{0.16\textwidth}
    >{\raggedright\arraybackslash}p{0.23\textwidth}X}
    \toprule
    \textbf{Dimension} & \textbf{Measure} & \textbf{Definition} \\
    \midrule
    Probability quality &
    Brier score; log loss &
    Unscaled three-class Brier score and negative log probability assigned to the realized 90-minute outcome. Analogous two-class scores are computed for advancement forecasts. \\

    Categorical quality &
    Modal H/D/A accuracy; advancement accuracy &
    Whether the realized class belongs to the set of maximum-probability classes. If $k$ classes share the maximum, each receives credit $1/k$. Advancement uses the larger of the two advancement probabilities. \\

    Scoreline quality &
    Exact score; goal difference; tendency; absolute errors &
    Indicators for the exact 90-minute score, correct signed goal difference, and correct H/D/A tendency implied by the scoreline. Absolute errors cover home goals, away goals, total goals, and goal difference. \\

    Game-style score &
    Scoring System points &
    Five points for the exact score, two for the correct goal difference, one for the correct tendency, and zero otherwise. The categories are mutually exclusive. \\

    Calibration &
    Outcome-specific reliability &
    For home, draw, and away separately, mean predicted probabilities are compared with observed frequencies in five approximately equal-frequency bins. Uncertainty resamples complete matches. \\

    Self reported confidence &
    Within-condition confidence association &
    Confidence is ranked within each model-version, information-access, and prompting-strategy condition. Brier score and modal accuracy are compared across relative confidence groups. \\

    Scoreline probability agreement &
    Modal agreement; mean-goal distance; sum audits &
    Agreement between the H/D/A outcome implied by the most likely scoreline and the outcome with the highest reported probability; distance between predicted mean goals and scoreline goals; deviations from required probability sums. \\

    Forecast diversity &
    Probability correlation; Jensen--Shannon divergence; same-condition ensemble &
    Pairwise similarity of model probability vectors within identical match and benchmark conditions, plus the performance of an equal-weight probability average formed without mixing conditions. \\

    Output validity &
    Validity, repair, normalization, and missingness rates &
    Shares of scheduled calls that are directly valid, valid after deterministic normalization, valid after one repair, invalid after repair, API errors, or timeouts. \\

    Observed search use &
    Search-observed and trace-availability rates &
    Among assigned open-book calls, the shares with observed search, no observed search, and unknown status. Closed-book calls form a separate tools-disabled audit. \\

    Resource use &
    Latency, input and output tokens, and cost &
    Per-attempt distributions and aggregate consumption by model version and benchmark condition. Failed calls remain in operational denominators when metadata are available. \\

    Generated rationale &
    Length and evidence-category prevalence &
    Word count and adjudicated mentions of prespecified evidence types in non-empty rationales. Categories describe generated text, not factual correctness, actual source use, or private model reasoning. \\

    Tournament forecasts &
    Choice scores, set recovery, and probability sums &
    Brier score, log loss, and final-selection accuracy for one-team questions; exact four-team set recovery, mean team recovery, marginal probability error, and probability-sum audit for semifinalists. \\
    \bottomrule
  \end{tabularx}
\end{table*}

\subsection{Statistical Analysis}

The primary match analysis evaluates T--24h H/D/A forecasts from the balanced seven-model panel without imputation. Information access is analyzed by assigned closed-book or open-book condition regardless of observed search. A secondary sensitivity analysis retains only open-book forecasts with observed search. Model-version and prompting-strategy comparisons hold the remaining benchmark conditions fixed.

Each paired contrast is reduced to one difference per match after the stated aggregation. We estimate uncertainty with 10,000 studentized bootstrap replicates that resample complete matches within group-stage and knockout strata. We test paired mean differences with 10,000 sign-flip permutations. We report estimates, $95\%$ intervals, raw $p$-values, Holm-adjusted $p$-values where applicable, and the relevant match and record counts. The declared Holm families cover information-access and prompting-strategy contrasts, complete-panel model pairs, operational snapshots, eligible stages, ensemble comparisons, and the soccer-specific external baseline. All random procedures use master seed 20260715.

Robustness checks preserve the paired design. They restrict realized lead times, repeat the information-access estimate after deleting each match in turn, summarize tournament stages, and resample complete matches for calibration intervals. Tournament forecasts are analyzed at the question level because all configurations for one question share the same realized outcome. Confidence and rationale results are associative; they do not establish causal information use or private model reasoning.

\subsection{Output Validation}

The archive contains 9,984 scheduled match attempts. The seven main model versions each provide a complete panel of 1,248 valid forecasts, giving the 8,736 forecasts used in the main evaluation. Claude Fable contributes 580 additional valid forecasts and 668 API errors. Across the full archive, 9,106 responses are directly valid and 210 become valid after one repair. No match response requires deterministic probability normalization. Table~\ref{tab:output-validity} reports the model-specific counts.

\begin{table*}[t]
  \centering
  \caption{Validation outcomes for all scheduled match attempts. Repaired responses are valid for scoring. Claude Fable is excluded from the balanced main analysis.}
  \label{tab:output-validity}
  \small
  \begin{tabular}{lrrrrr}
    \toprule
    \textbf{Model version} & \textbf{Scheduled} & \textbf{Directly valid} & \textbf{Valid after repair} & \textbf{API error} & \textbf{Scorable} \\
    \midrule
    GPT 5.5 & 1,248 & 1,248 & 0 & 0 & 1,248 \\
    Claude Opus 4.8 & 1,248 & 1,246 & 2 & 0 & 1,248 \\
    Gemini 3.1 Pro Preview & 1,248 & 1,194 & 54 & 0 & 1,248 \\
    Grok 4.3 & 1,248 & 1,246 & 2 & 0 & 1,248 \\
    DeepSeek V4 Pro & 1,248 & 1,230 & 18 & 0 & 1,248 \\
    Qwen 3.7 Max & 1,248 & 1,232 & 16 & 0 & 1,248 \\
    Mistral Large 2512 & 1,248 & 1,134 & 114 & 0 & 1,248 \\
    \midrule
    Main panel & 8,736 & 8,530 & 206 & 0 & 8,736 \\
    Claude Fable 5 & 1,248 & 576 & 4 & 668 & 580 \\
    \midrule
    Full archive & 9,984 & 9,106 & 210 & 668 & 9,316 \\
    \bottomrule
  \end{tabular}
\end{table*}

\subsection{Artifact Provenance}

Every figure and table is accompanied by a machine-readable JSON artifact containing its estimand, estimate, interval, $p$-values, counts, aggregation rule, configuration hash, source-table hashes, output hash, and manifest key. The manifest also records the Python, R, package, operating-system, and external-tool versions.

The final manifest links the complete database SHA-256 to every derived table and paper artifact. Acceptance checks cover database integrity, the 104-match outcome universe, stable keys, probability bounds and sums, metric recomputation, and the complete pre-results checklist. The public export contains the same 9,984 match-attempt rows and reproduces the stored probabilities, Brier scores, and log losses.

\subsection{Public Platform Views}

\begin{figure*}[t]
  \centering
  \begin{subfigure}[t]{0.485\textwidth}
    \centering
    \includegraphics[width=\linewidth]{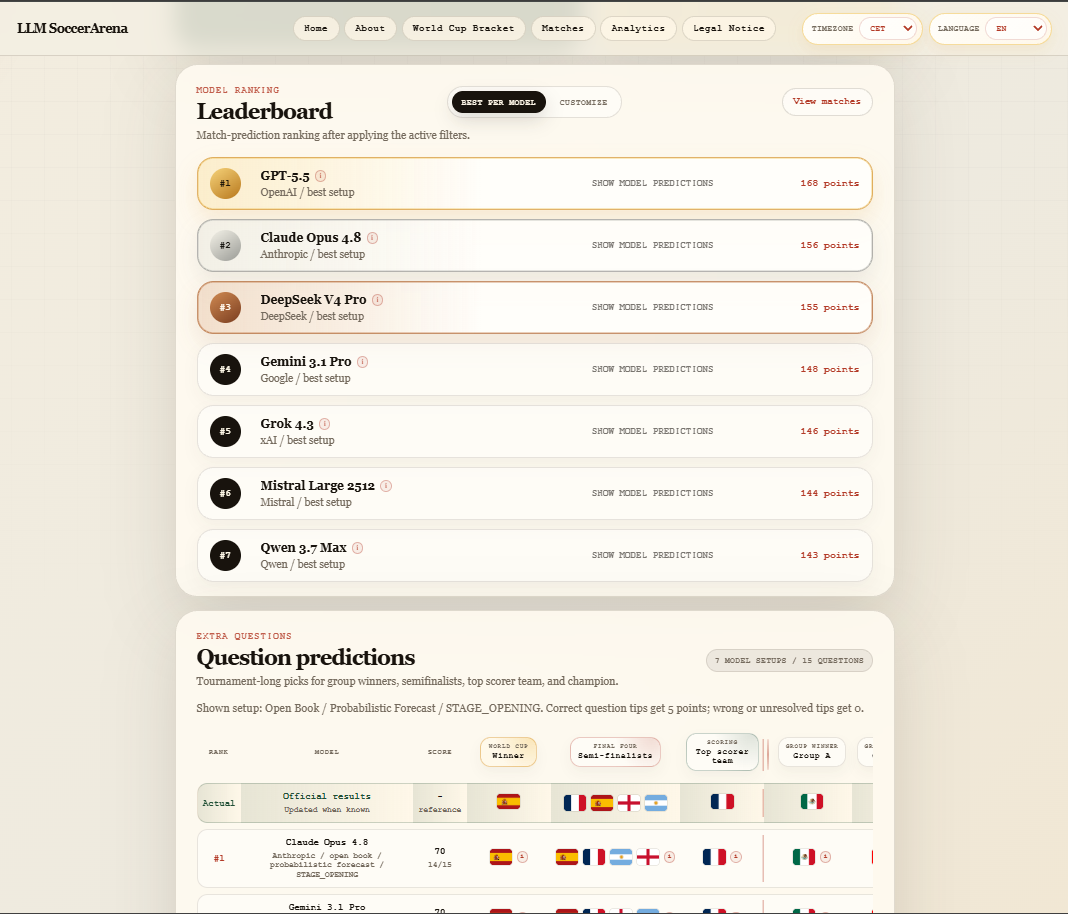}
    \caption{Match forecasts with scorelines, probabilities, benchmark conditions, and generated rationales.}
  \end{subfigure}\hfill
  \begin{subfigure}[t]{0.485\textwidth}
    \centering
    \includegraphics[width=\linewidth]{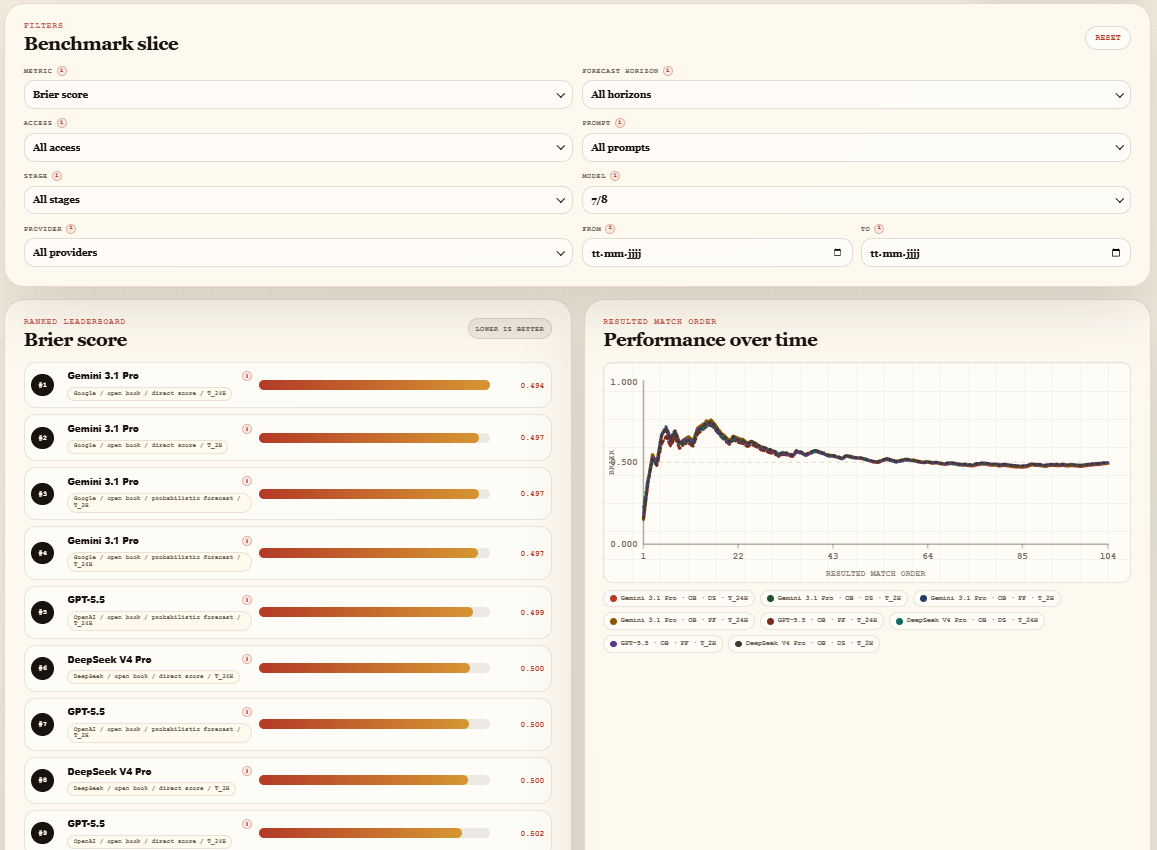}
    \caption{Filterable performance summaries and trajectories.}
  \end{subfigure}
  \caption{Public interfaces to the auditable benchmark archive. Figure~\ref{fig:platform} summarizes the platform architecture; this figure shows the records that readers can inspect on the public website.}
  \label{fig:website}
\end{figure*}

\onecolumn
\raggedbottom

\section{Supplementary Results}

\subsection{Results}
\label{sec:results}

\begin{multicols}{2}
We organize the results around five evaluation dimensions: \textbf{(1) model performance}, whether model versions differ in forecast quality and provide distinct forecast signals; \textbf{(2) information access}, whether current web information is retrieved, changes forecasts, and improves accuracy; \textbf{(3) prompting strategy}, whether score-first or probabilistic forecast prompting changes forecast quality and cross-field consistency; \textbf{(4) calibration}, whether reported probabilities and self-reported confidence correspond to observed accuracy; and \textbf{(5) tournament forecasts}, whether LLMs can produce coherent longer-horizon predictions about the tournament structure. Unless stated otherwise, match results use the balanced seven-model panel and compare forecasts on the same matches under matched benchmark conditions.

The supplementary results follow the five evaluation dimensions in the main paper. They provide the robustness checks, diagnostic analyses, and complete figures that support the concise main-text findings. The final subsection reports a soccer-specific closing-odds comparison. This external baseline is deliberately confined to the appendix because it is not part of the general \arena protocol.

\subsection{Model Performance Details}
\label{app:performance-details}

\textbf{Model comparisons.}
The complete family contains 21 paired Brier-score comparisons among the seven model versions. None remains significant after Holm correction. The smallest adjusted $p$-value is $0.055$ for GPT compared with Mistral. The descriptive leaderboard therefore does not establish statistically distinct model tiers.

\textbf{Tournament progression.}
Cumulative Brier score and modal accuracy are volatile early and stabilize as matches accumulate. Match-level median Brier increases from $0.403$ in the group stage to $0.453$ in the round of 32, $0.504$ in the round of 16, $0.591$ in the quarterfinals, and $0.663$ in the semifinals. The late stages contain few matches, so this pattern describes the realized tournament rather than a general increase in difficulty.
\end{multicols}

\begin{figure}[H]
  \centering
  \includegraphics[width=\textwidth,height=0.78\textheight,keepaspectratio]{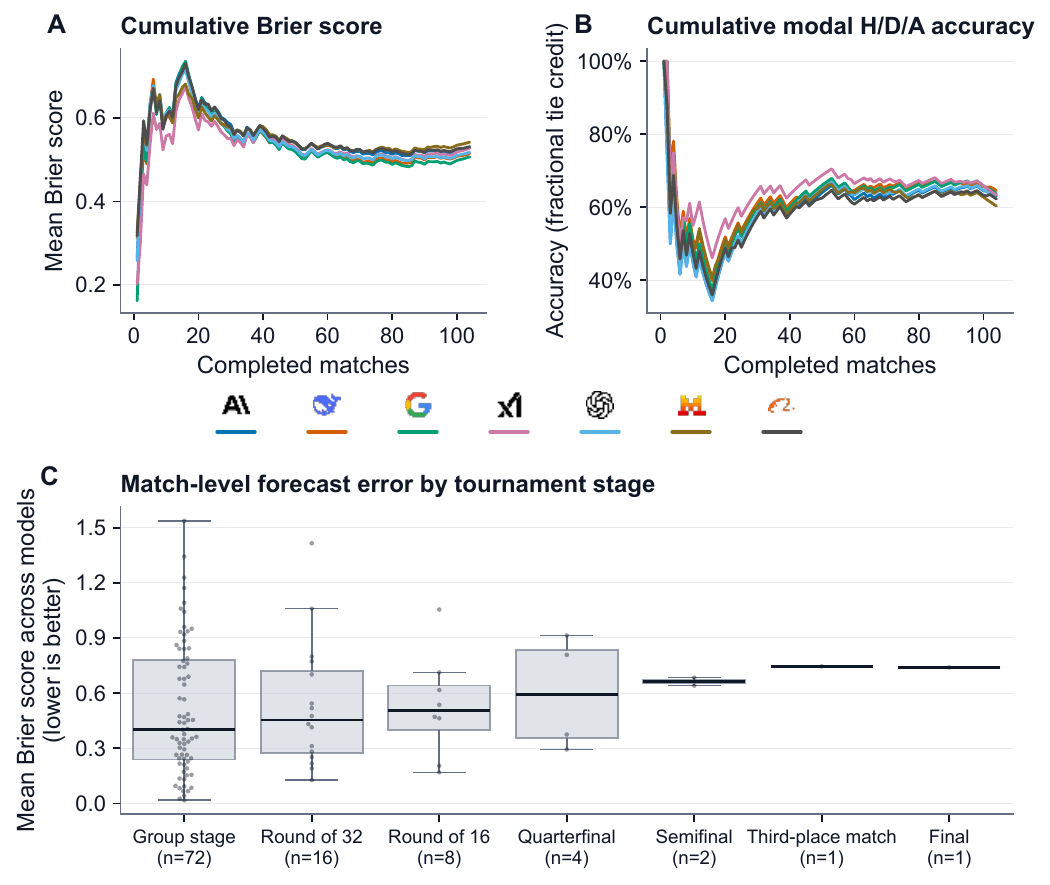}
  \caption{Performance across the tournament. Cumulative trajectories use the balanced T--24h aggregation across both information-access conditions and both prompting strategies. Stage distributions show one complete-panel mean per match with raw observations and stage sample sizes.}
  \label{fig:overall}
\end{figure}

\begin{multicols}{2}
\textbf{Forecast similarity.}
We compare model versions only within the same T--24h match, information-access condition, and prompting strategy. Mean pairwise Jensen--Shannon divergence is $0.00442$ nats, with a range from $0.00162$ to $0.00834$. Mean pairwise probability correlation is $0.943$, with a range from $0.903$ to $0.984$. These values show that the seven model versions produce highly similar probability forecasts.

Open-book access does not detectably increase this convergence. Mean pairwise Jensen--Shannon divergence is $0.00452$ in the closed-book condition and $0.00432$ in the open-book condition. The paired open-minus-closed difference is $-0.2\times10^{-3}$ nats, with a $95\%$ confidence interval of $[-0.9,0.6]\times10^{-3}$ and $p=0.593$.

\textbf{Same-condition ensemble.}
We average the seven probability vectors only within the same match and benchmark conditions. The ensemble Brier score is $0.0047$ lower than the average member score. Under Brier loss, this difference is the nonnegative disagreement term created by averaging nonidentical probability forecasts \citep{krogh_vedelsby_1994}. The ensemble directionally outperforms four model versions and trails three; none of the seven ensemble comparisons remains significant after Holm correction.
\end{multicols}

\begin{figure}[H]
  \centering
  \includegraphics[width=\textwidth,height=0.78\textheight,keepaspectratio]{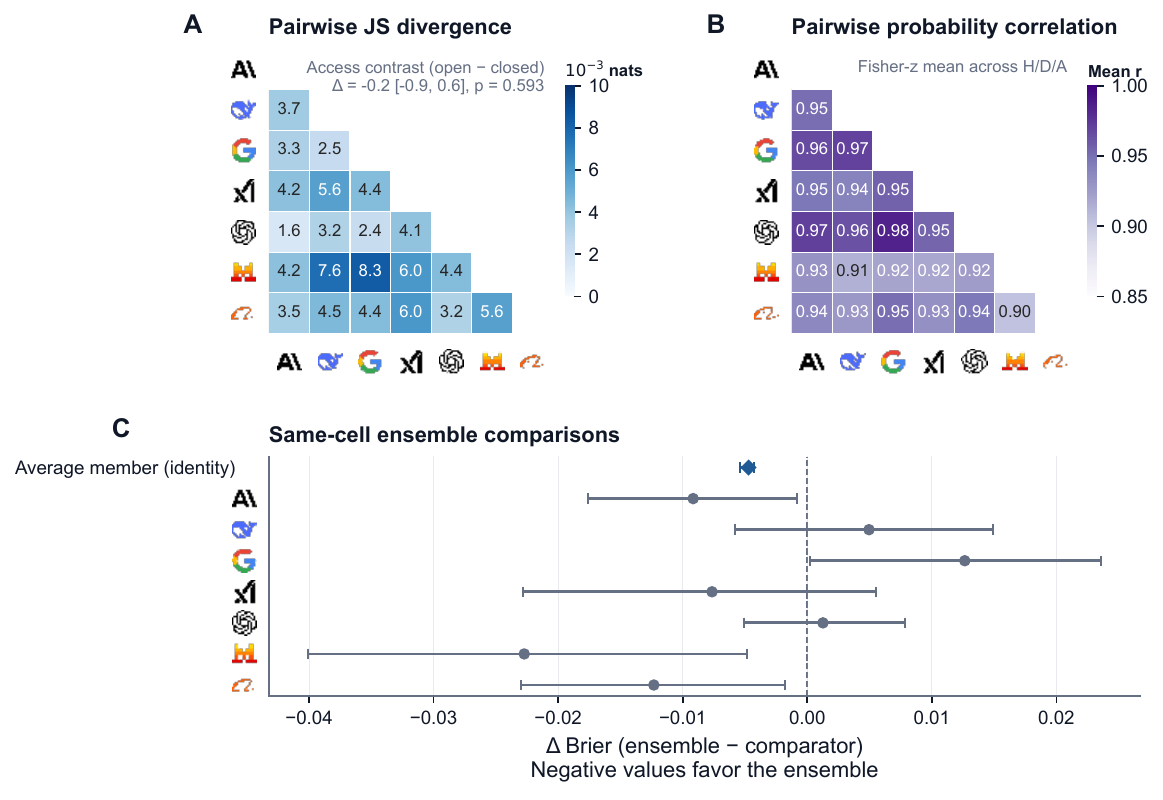}
  \caption{Forecast similarity and same-condition ensemble performance. The matrices report pairwise Jensen--Shannon divergence and combined probability correlation. The forest plot compares the equal-weight ensemble with the average member and each model version without mixing forecast horizons, information-access conditions, or prompting strategies.}
  \label{fig:forecast-diversity}
  \label{fig:rq4}
\end{figure}

\subsection{Information Access Details}
\label{app:access-details}

\begin{multicols}{2}
\textbf{Robustness.}
The T--24h closed-minus-open Brier improvement remains stable when any one match is removed. The leave-one-match-out estimates range from $0.0208$ to $0.0253$. Restricting the analysis to calls made within $1{,}440\pm90$ minutes of kickoff gives an effect of $0.0217$, with a $95\%$ confidence interval of $[0.0021,0.0401]$, across 101 matches. The three excluded matches fall outside this prespecified timing window. The primary analysis retains all 104 matches.

The effect is largest in the round of 32, where the paired difference is $0.0611$, with a $95\%$ confidence interval of $[0.0216,0.0935]$ and a Holm-adjusted $p$-value of $0.012$. Stage-specific estimates remain secondary because the knockout stages contain few matches.

\textbf{Observed search sensitivity.}
The primary analysis compares assigned open-book and closed-book conditions. This intent-to-treat comparison preserves the factorial design even when an open-book model does not search. A secondary sensitivity analysis restricts open-book forecasts to calls with observed search. It measures the association with actual search use but no longer compares the original randomized conditions and can reflect model-specific search behavior.

\begin{figure}[H]
  \centering
  \includegraphics[width=0.82\textwidth,height=0.72\textheight,keepaspectratio]{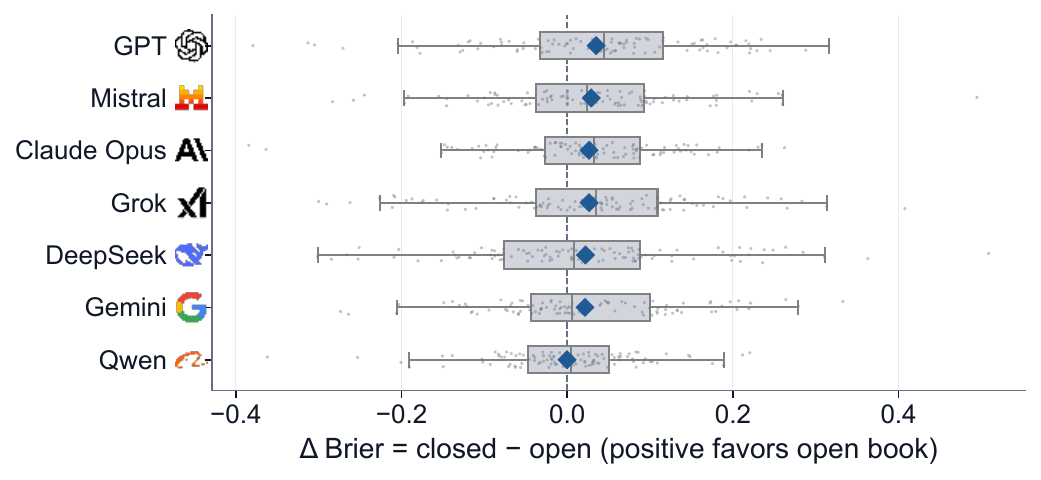}
  \caption{Model-specific distributions of T--24h closed-minus-open Brier-score differences. The pooled access contrast is prespecified; model-specific results are descriptive.}
  \label{fig:access-by-model}
  \label{fig:rq1_access_by_model}
\end{figure}

\textbf{Model-specific effects.}
Access gains vary descriptively across model versions. They range from approximately zero for Qwen to $0.0349$ for GPT. These estimates are not treated as separate confirmatory effects. The model-specific distributions appear in Figure~\ref{fig:access-by-model}.

\textbf{Variation across factors.}
A two-way decomposition of the $7\times4$ matrix of T--24h cell means attributes $44.8\%$ of the variation to the four information-access and prompting-strategy conditions, $40.5\%$ to model version, and $14.7\%$ to their interaction. Bootstrap distributions are broad. The decomposition supports treating information access as a first-class benchmark factor without claiming that it dominates every model difference.
\end{multicols}

\begin{figure}[H]
  \centering
  \includegraphics[width=\textwidth,height=0.78\textheight,keepaspectratio]{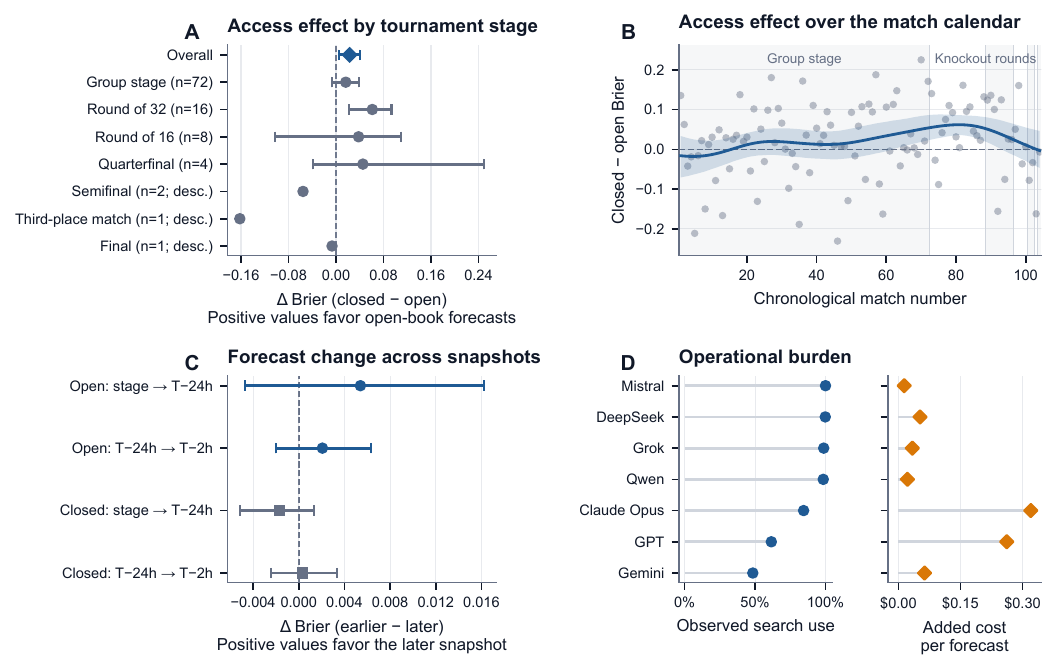}
  \caption{Information-access results. Panels show the pooled T--24h access effect, horizon comparisons, forecast changes across snapshots, and operational burden. Positive Brier differences favor open-book forecasts.}
  \label{fig:access-details}
\end{figure}

\subsection{Prompting Strategy Details}
\label{app:prompting-details}

\begin{multicols}{2}
\textbf{Complete metric comparison.}
We orient effects so that positive values favor probabilistic forecast prompting. The Brier-score effect is $-0.0008$, with a $95\%$ confidence interval of $[-0.0044,0.0029]$ and a Holm-adjusted $p$-value of $0.693$. No significant effect appears for log loss, modal H/D/A accuracy, exact-score accuracy, or Scoring System points. Modal accuracy directionally favors score-first prompting by $0.89$ percentage points, but the difference does not remain significant after correction.

\textbf{Information-access interaction.}
Score-first prompting is directionally better in the closed-book condition, while probabilistic forecast prompting is directionally better in the open-book condition. The additional probabilistic forecast advantage under open-book access is $0.0067$, with a $95\%$ confidence interval of $[0.0001,0.0138]$, raw $p=0.055$, and Holm-adjusted $p=0.110$.

\textbf{Reported forecasts.}
Probabilistic forecast prompting produces $3.98$ percentage points more score-implied draws, $4.05$ points more exact $1$--$1$ predictions, and $5.36$ points more cases in which the scoreline implies a draw while another H/D/A outcome has the highest probability. At the same time, the mean reported draw probability decreases by $0.21$ percentage points. Six of the seven model versions show the increase in draw scorelines.

\textbf{Scoreline probability agreement.}
Agreement means that the most likely scoreline implies the outcome with the highest H/D/A probability, such as $2{:}1$ together with home win. In the closed-book condition, agreement decreases from $69.0\%$ under score-first prompting to $62.4\%$ under probabilistic forecast prompting. In the open-book condition, it decreases from $83.8\%$ to $79.7\%$. Disagreement does not necessarily make a forecast invalid because one exact scoreline can be most likely even when another aggregate outcome has the highest probability.

\textbf{Output validation.}
No prompting-strategy condition requires deterministic probability normalization. Repairs are retained as valid forecasts when the repaired response satisfies the original schema. Table~\ref{tab:output-validity} reports the complete match-level validation counts.
\end{multicols}

\begin{figure}[H]
  \centering
  \includegraphics[width=\textwidth,height=0.76\textheight,keepaspectratio]{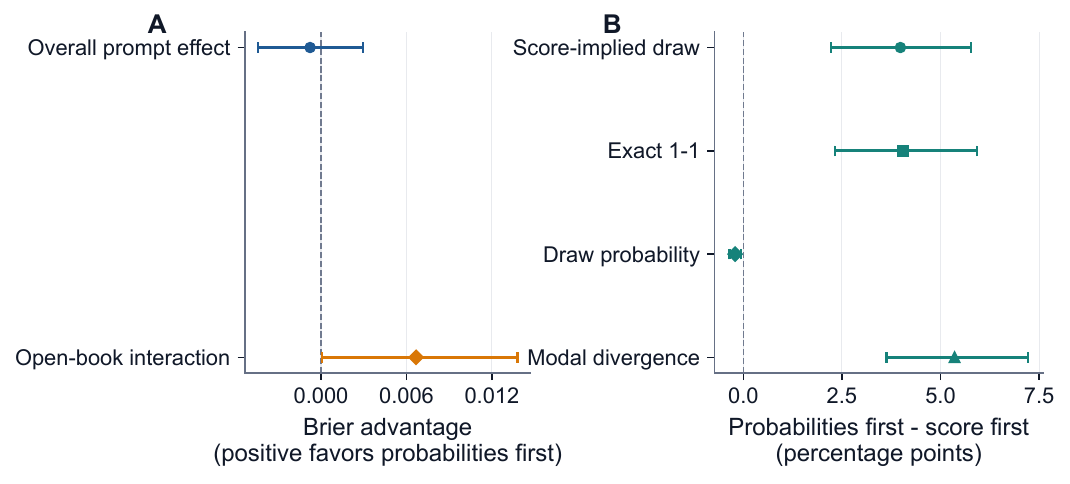}
  \caption{Prompting-strategy effects. The left panel reports paired Brier contrasts. The right panel reports changes in score-implied draws, exact $1$--$1$ predictions, mean draw probability, and scoreline-probability disagreement. Positive values favor probabilistic forecast prompting.}
  \label{fig:prompting-strategy}
\end{figure}

\begin{figure}[H]
  \centering
  \includegraphics[width=0.60\textwidth,height=0.54\textheight,keepaspectratio]{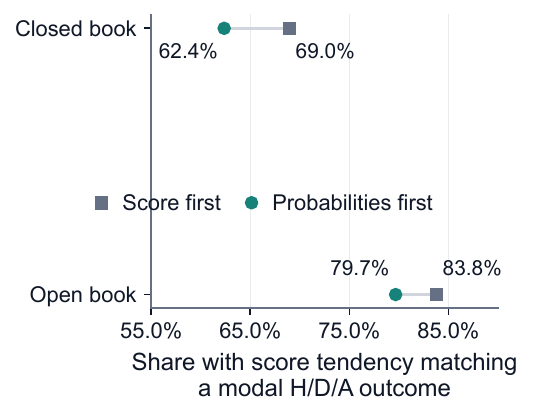}
  \caption{Agreement between the outcome implied by the most likely scoreline and the outcome with the highest reported probability. Exact-score and aggregate-outcome modes need not coincide, so disagreement is not labeled a logical inconsistency.}
  \label{fig:prompting-agreement}
  \label{fig:prompting-coherence}
\end{figure}

\subsection{Calibration Details}
\label{app:calibration-details}

\begin{multicols}{2}
\textbf{Probability calibration.}
For each T--24h match, information-access condition, and prompting strategy, we average the seven model probability vectors. This produces one probability vector for each of the $104\times2\times2=416$ matched conditions. We divide the predicted probabilities into five approximately equal-frequency bins separately for home, draw, and away outcomes. Pointwise intervals resample complete matches.

The calibration curves broadly follow the ideal diagonal, but several bins deviate and the uncertainty intervals are wide. Home probabilities vary around the diagonal without a simple directional pattern. Mean draw probability is $24.7\%$, compared with an observed draw rate of $27.9\%$. The predicted-minus-observed difference is $-3.2$ percentage points, with a $95\%$ confidence interval of $[-12.5,4.9]$ and $p=0.461$. The outcome-specific deviations do not establish one consistent bias across all three outcomes.

\textbf{Self reported confidence.}
We rank confidence within each model version, information-access condition, and prompting strategy because model versions use different confidence scales. We preserve ties and group forecasts by relative confidence rank. From the lowest to the highest group, mean Brier score decreases from $0.620$ to $0.423$, while modal H/D/A accuracy increases from $49.8\%$ to $73.1\%$. Confidence therefore identifies easier and harder forecasts within a model configuration, but it is not interpreted as a calibrated probability of correctness.
\end{multicols}

\begin{figure}[H]
  \centering
  \includegraphics[width=\textwidth,height=0.78\textheight,keepaspectratio]{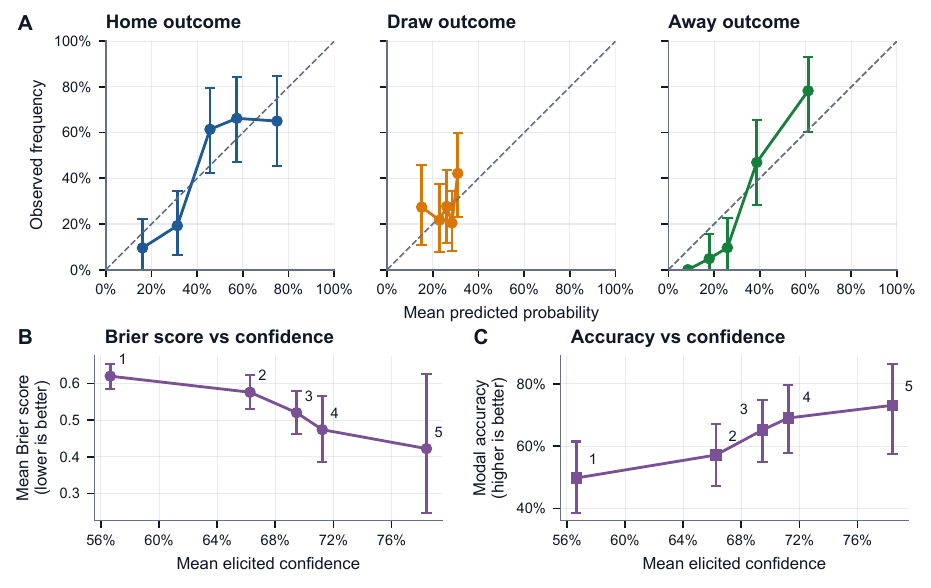}
  \caption{Probability calibration and self reported confidence. Home, draw, and away outcomes use separate reliability panels with pointwise match-bootstrap intervals. Confidence groups are defined within model version, information access, and prompting strategy; lower Brier score and higher modal accuracy indicate better forecasts.}
  \label{fig:calibration}
\end{figure}

\begin{multicols}{2}
\subsection{Tournament Forecast}
\label{app:tournament-details}

Before the tournament, the seven model versions answered 15 longer-horizon questions: the 12 group winners, the four semifinalists, the world champion, and the team of the top scorer. We evaluate both the final selections and the reported probabilities. Appendix~\ref{app:prompts}, including Table~\ref{tab:tournament-candidates}, provides the exact question definitions.

\textbf{Group winner.}
Across the 28 configurations ($7$ model versions $\times$ $2$ information access conditions $\times$ $2$ prompting strategies), $91.1\%$ of the 12 group winner selections are correct. However, all 28 select Portugal for Group K, which Colombia wins. This shared error shows that agreement does not guarantee correctness.

The top scorer team is also predicted well: 25 of 28 forecasts select France. The world champion is apparently much harder. Only 5 of 28 forecasts select Spain, while 18 select France. Figure~\ref{fig:combined-results}\textbf{(c)} summarizes the final selections across all questions.

\textbf{Semifinalists.}
Ten of 28 forecasts recover the exact four-team semifinal set. On average, each forecast selects $3.29$ of the four correct teams. The final selections are therefore often close even when the complete set is not correct.

The reported marginal probabilities show a separate problem. They should sum to four because exactly four teams reach the semifinals, but several forecasts violate this requirement. We therefore evaluate final set recovery separately from probability quality. Figure~\ref{fig:tournament-probability-sums} in Appendix~\ref{app:tournament-details} shows the complete probability sum audit across all 28 configurations.

\textbf{Question set.}
The seven model versions answer 15 questions once at stage opening under both information-access conditions and both prompting strategies. Each question therefore has 28 configurations. The questions cover the 12 group winners, the four semifinalists, the world champion, and the team of the tournament's top scorer. Appendix~\ref{app:prompts} provides the exact question definitions and candidate lists.

\textbf{One-team questions.}
Across the 12 group-winner questions, $91.1\%$ of final selections are correct. Group A has 26 correct selections, ten groups have 28, and Group K has none. All 28 configurations select Portugal for Group K, which Colombia wins. For the two tournament-wide one-team questions, 25 configurations select France as the team of the top scorer and five select Spain as champion. Figure~\ref{fig:combined-results}\textbf{(c)} reports the probability assigned to each realized outcome and the correct-selection counts.

\textbf{Semifinalists.}
Ten of 28 configurations recover the exact four-team semifinal set. The mean forecast contains $3.29$ of the four correct teams. We evaluate final-set recovery separately from the marginal inclusion probabilities because the latter must satisfy an additional sum constraint.

\textbf{Probability-sum audit.}
The 48 marginal semifinal probabilities should sum to four because exactly four teams reach the semifinals. Several configurations violate this requirement, with strongly model-specific deviations. We audit the reported probabilities without retrospective normalization. Figure~\ref{fig:tournament-probability-sums} shows the complete distribution.
\end{multicols}

\begin{figure}[H]
  \centering
  \includegraphics[width=\textwidth,height=0.78\textheight,keepaspectratio]{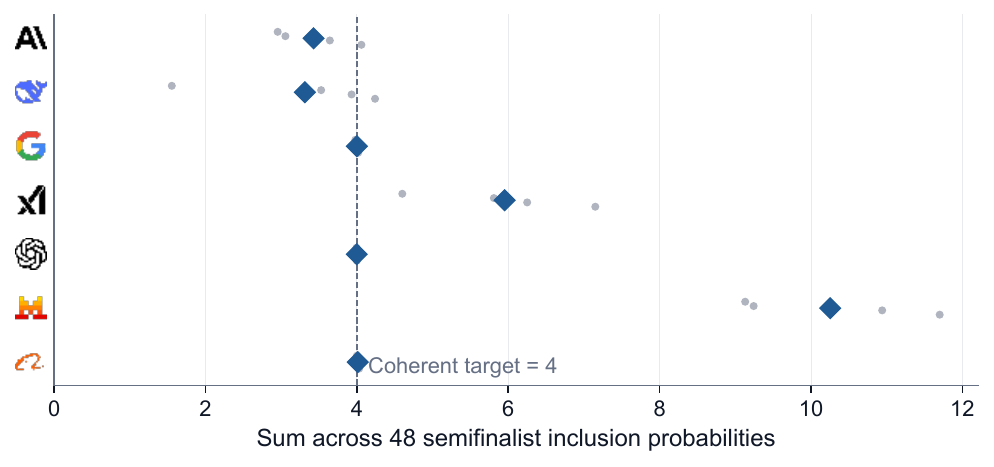}
  \caption{Sum of the 48 reported semifinal inclusion probabilities for each model version and benchmark condition. The required sum is four. Raw marginals are audited without retrospective normalization.}
  \label{fig:tournament-probability-sums}
  \label{fig:rq6_probability_sums}
\end{figure}

\section{Rationale Analysis}
\label{app:rationale-analysis}

\begin{multicols}{2}
\subsection{Scope}

The rationale analysis covers 1,456 complete-panel T--24h probabilistic forecast rationales. It studies generated text and observable provider traces, not private model reasoning. In the balanced seven-model panel, $84.5\%$ of open-book forecasts show observed search. Model-specific rates range from $48.4\%$ to $100\%$. Closed-book search is not applicable because tools are disabled by design.

\subsection{Annotation Procedure}

A released keyword lexicon and the frozen annotator \texttt{z-ai/glm-5.2} label every rationale independently. The annotator receives a fixed prompt, a temperature of zero, and no access to outcomes, forecast performance, model identity, or keyword labels. Raw responses from 73 batches covering all 1,456 rationales are cached and frozen because temperature zero does not guarantee identical regeneration.

The prespecified categories are markets or odds, recent form, injuries or lineups, rankings or strength, tactics, venue or travel, tournament context, explicit sources, and generic unsupported claims. The labels indicate that a rationale mentions a category. They do not establish factual correctness, actual source use, causal influence on the forecast, or private reasoning.

\subsection{Human Audit}

One blinded human auditor labels a fixed-seed sample of 14 rationales per model-version and information-access cell, giving 196 texts. Final prevalence estimates use human labels for these 196 texts and frozen GLM labels for the remaining 1,260. The keyword coder remains an independent comparison.

Across the full scoped corpus, keyword-GLM agreement ranges from $48.0\%$ to $92.7\%$, with $\kappa$ from $0.04$ to $0.75$. Against the human audit, GLM reaches $89.8\%$ to $100\%$ agreement and $\kappa=0.59$ to $1.00$ for seven categories. Agreement is weaker for tournament context at $57.1\%$ and $\kappa=0.25$. Generic unsupported claims reach $83.2\%$ agreement but $\kappa=0.10$ because positive cases are rare. Human-only information-access differences have the same direction as the resolved estimates for every category. Tournament context is therefore retained with this validation caveat.

\subsection{Complete Evidence Categories}

The main-text evidence directions hold for every model version, although their size differs. GPT shows the richest current-evidence profile. Qwen searches frequently but shows smaller open-book increases in odds, injuries, and explicit sources. Grok produces the shortest rationales and the largest reduction in generic unsupported claims. These differences describe generated text and do not rank reasoning quality.
\end{multicols}

\begin{figure}[H]
  \centering
  \includegraphics[width=\textwidth,height=0.78\textheight,keepaspectratio]{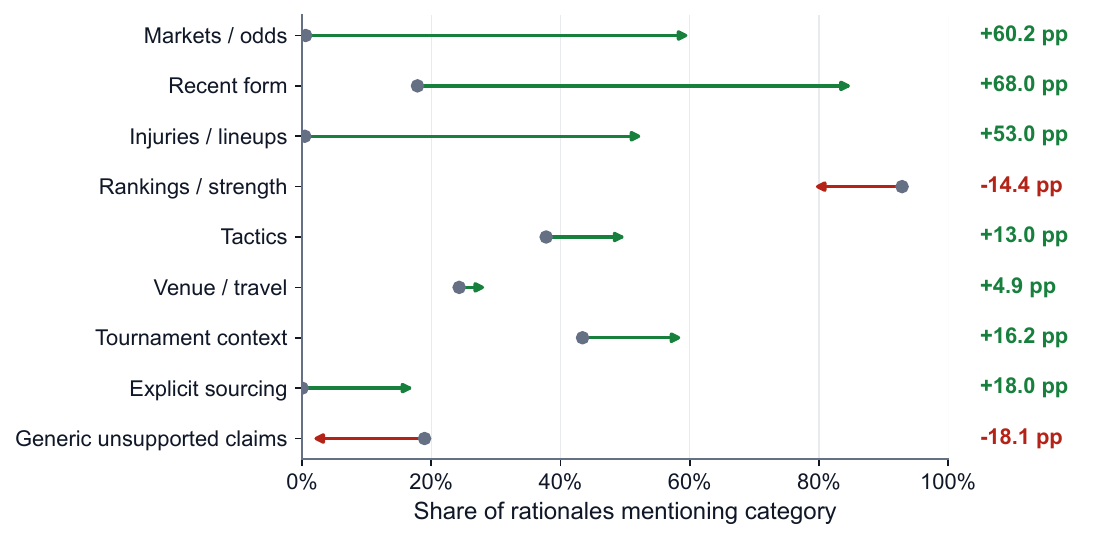}
  \caption{Complete evidence-category results for the 1,456 T--24h probabilistic forecast rationales. Points mark closed-book prevalence and arrows terminate at open-book prevalence. Tournament context has weaker GLM-human validation and should be read with the stated annotation sensitivity.}
  \label{fig:rationale-evidence-complete}
  \label{fig:rq5_evidence_appendix}
\end{figure}

\begin{figure}[H]
  \centering
  \includegraphics[width=\textwidth,height=0.78\textheight,keepaspectratio]{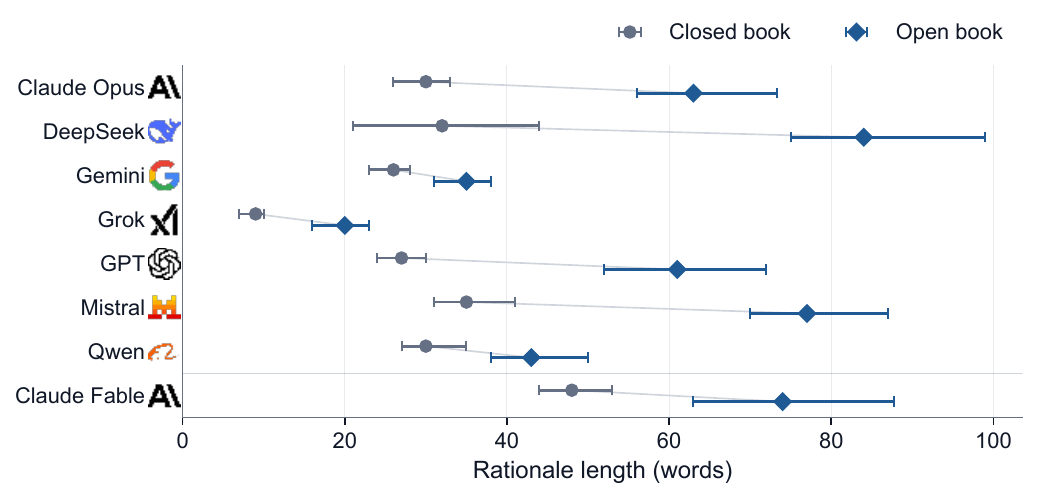}
  \caption{Model-specific median and interquartile range of rationale length by information-access condition. Claude Fable appears only as additional archive coverage: 580 of its 1,248 planned match cells contain a valid rationale, while the seven main model versions cover every planned cell.}
  \label{fig:rationale-length}
  \label{fig:rq5_length}
\end{figure}

\begin{figure}[H]
  \centering
  \includegraphics[width=\textwidth,height=0.78\textheight,keepaspectratio]{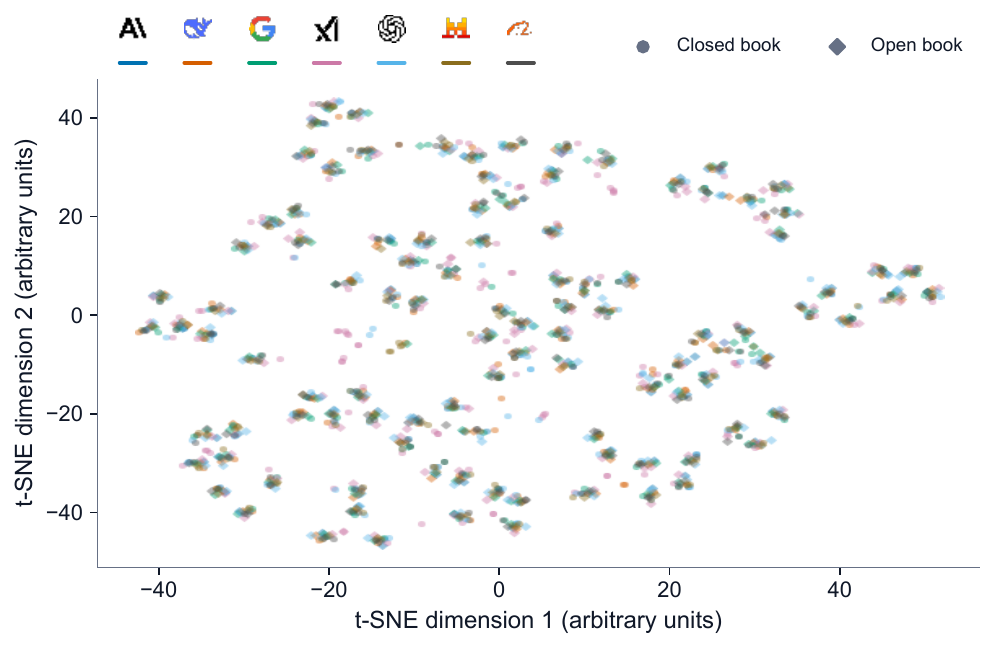}
  \caption{Exploratory fixed-seed t-SNE projection of frozen, L2-normalized MPNet embeddings for the 1,456 rationale texts. Color denotes model version and marker denotes information access. Axes and distances are arbitrary; the projection supports no confirmatory or mechanistic interpretation.}
  \label{fig:rationale-tsne}
  \label{fig:rq5_tsne_appendix}
\end{figure}

\section{Soccer-specific Closing Odds Comparison}
\label{app:odds-comparison}

\begin{multicols}{2}
This analysis compares \arena with a soccer-specific external reference. It is included only in the appendix because closing odds are not part of the general benchmark protocol and do not transfer directly to other domains.

\subsection{Source and Timing}

De-vigged bookmaker consensus is an established forecasting reference because it aggregates information and judgments across market participants \citep{leitner_etal_2010,zeileis_etal_2018}. We retrieve odds from the historical endpoint of The Odds API for the \texttt{soccer\_fifa\_world\_cup} sport key, European bookmaker region, and \texttt{h2h} market. Prices are decimal odds for the 90-minute home, draw, and away outcomes.

For each distinct kickoff time, the collector requests the closest historical snapshot at or before one second before kickoff. Events are matched to the local fixture by kickoff time within one minute and by oriented team names. The extract covers all 104 matches and contains 2,487 bookmaker-match rows, with 15 to 25 bookmakers per match and a median of 24. Each row stores the provider event ID, bookmaker, requested and returned snapshot times, market update time, oriented fixture key, and three decimal prices. The 92 distinct raw snapshots are cached and hash verified. Because the feed is subscription data, raw odds remain local; derived consensus probabilities and results are released.

\subsection{De-vigging and Consensus}

For match $m$, bookmaker $b$, and outcome $k\in\{H,D,A\}$, let $o_{mbk}>1$ denote the decimal odds. Raw inverse odds sum to more than one when the bookmaker margin is positive. We remove the margin within each bookmaker:
\begin{equation}
  p_{mbk}^{\mathrm{fair}}
  = \frac{1/o_{mbk}}{\sum_{j\in\{H,D,A\}} 1/o_{mbj}}.
  \label{eq:devig}
\end{equation}
We then take the cross-bookmaker median for each outcome and renormalize the three values:
\begin{align}
  \widetilde p_{mk}
  &= \operatorname{median}_{b}\!\left(p_{mbk}^{\mathrm{fair}}\right), \\
  p_{mk}^{\mathrm{market}}
  &= \frac{\widetilde p_{mk}}{\sum_{j\in\{H,D,A\}}\widetilde p_{mj}}.
  \label{eq:market-consensus}
\end{align}
This yields one valid closing-odds probability vector per match without selecting a preferred bookmaker or imputing prices.

\subsection{Matched Comparison}

We compare closing odds with the balanced seven-model open-book, probabilistic forecast T--2h forecasts. This is the closest registered LLM condition, although the market can still incorporate information arriving during the final two hours. Table~\ref{tab:closing-odds-comparison} reports the seven model versions, their equal-weight probability ensemble, and the market reference.
\end{multicols}

\begin{table}[H]
  \centering
  \caption{Closing odds versus balanced open-book, probabilistic forecast LLM forecasts at T--2h over all 104 matches. Values are means with $95\%$ stratified match-bootstrap intervals. Brier skill is $100\times(1-\mathrm{Brier}_{\mathrm{LLM}}/\mathrm{Brier}_{\mathrm{market}})$, so positive values favor the LLM. Lower Brier score, log loss, and ranked probability score are better; higher modal H/D/A accuracy is better.}
  \label{tab:closing-odds-comparison}
  \resizebox{\textwidth}{!}{%
  \begin{tabular}{llrlllr}
  \toprule
  Forecaster & Brier [95\% CI] & Brier skill vs. market, \% & Log loss [95\% CI] & RPS [95\% CI] & Modal H/D/A accuracy, \% [95\% CI] & $n$ \\
  \midrule
  Gemini & 0.497 [0.423, 0.582] & 0.3 & 0.842 [0.738, 0.958] & 0.150 [0.128, 0.177] & 63.5 [54.0, 72.7] & 104 \\
  Closing odds & 0.498 [0.424, 0.584] & 0.0 & 0.844 [0.741, 0.958] & 0.150 [0.128, 0.176] & 63.5 [54.1, 72.7] & 104 \\
  GPT & 0.500 [0.429, 0.581] & $-0.3$ & 0.846 [0.749, 0.953] & 0.152 [0.131, 0.176] & 64.4 [54.6, 73.1] & 104 \\
  DeepSeek & 0.506 [0.434, 0.586] & $-1.5$ & 0.859 [0.761, 0.972] & 0.153 [0.133, 0.178] & 63.5 [54.0, 72.7] & 104 \\
  LLM ensemble & 0.507 [0.439, 0.585] & $-1.7$ & 0.857 [0.765, 0.962] & 0.154 [0.134, 0.176] & 64.4 [54.9, 73.7] & 104 \\
  Grok & 0.511 [0.437, 0.598] & $-2.6$ & 0.865 [0.759, 0.989] & 0.153 [0.132, 0.179] & 62.5 [52.2, 71.8] & 104 \\
  Claude Opus & 0.517 [0.451, 0.592] & $-3.8$ & 0.871 [0.779, 0.973] & 0.158 [0.139, 0.180] & 61.5 [51.2, 70.9] & 104 \\
  Mistral & 0.518 [0.457, 0.586] & $-4.0$ & 0.875 [0.794, 0.969] & 0.159 [0.142, 0.179] & 65.9 [56.3, 74.6] & 104 \\
  Qwen & 0.529 [0.462, 0.601] & $-6.2$ & 0.888 [0.797, 0.989] & 0.163 [0.144, 0.185] & 60.1 [50.2, 69.6] & 104 \\
  \bottomrule
  \end{tabular}}
\end{table}

\begin{multicols}{2}
Gemini is descriptively closest to the market, with Brier score $0.497$ compared with $0.498$ for closing odds. Its paired market-minus-LLM advantage is $0.001$, with a $95\%$ confidence interval of $[-0.006,0.008]$ and a Holm-adjusted $p$-value of $1.000$. One of the seven individual model versions has a positive mean Brier advantage over closing odds, and none remains positive after the prespecified correction. These intervals describe uncertainty within this tournament, not performance across future tournaments.
\end{multicols}

\begin{figure}[H]
  \centering
  \includegraphics[width=\textwidth,height=0.78\textheight,keepaspectratio]{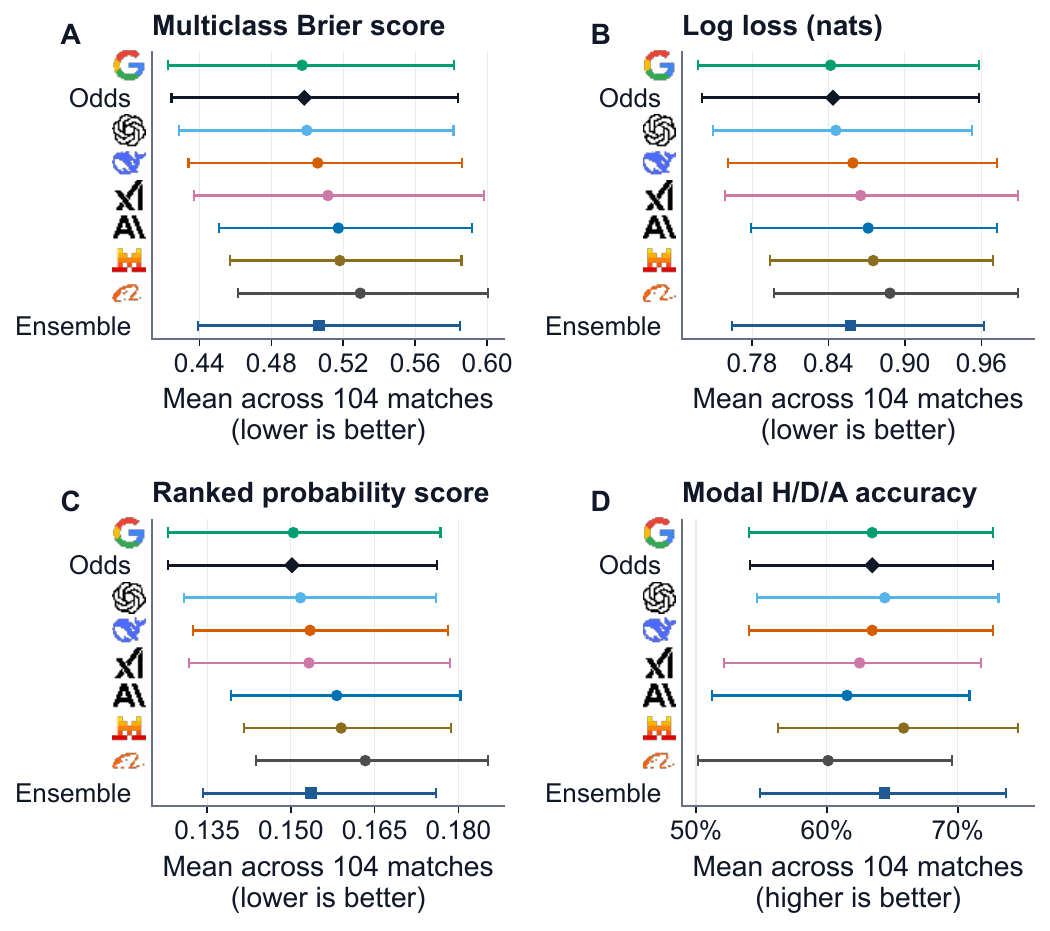}
  \caption{Absolute performance of closing odds, individual model versions, and the equal-weight LLM ensemble in the matched open-book, probabilistic forecast T--2h condition. Points are means across 104 matches; bars are $95\%$ stratified match-bootstrap intervals.}
  \label{fig:closing-odds-absolute}
\end{figure}

\begin{figure}[H]
  \centering
  \includegraphics[width=\textwidth,height=0.78\textheight,keepaspectratio]{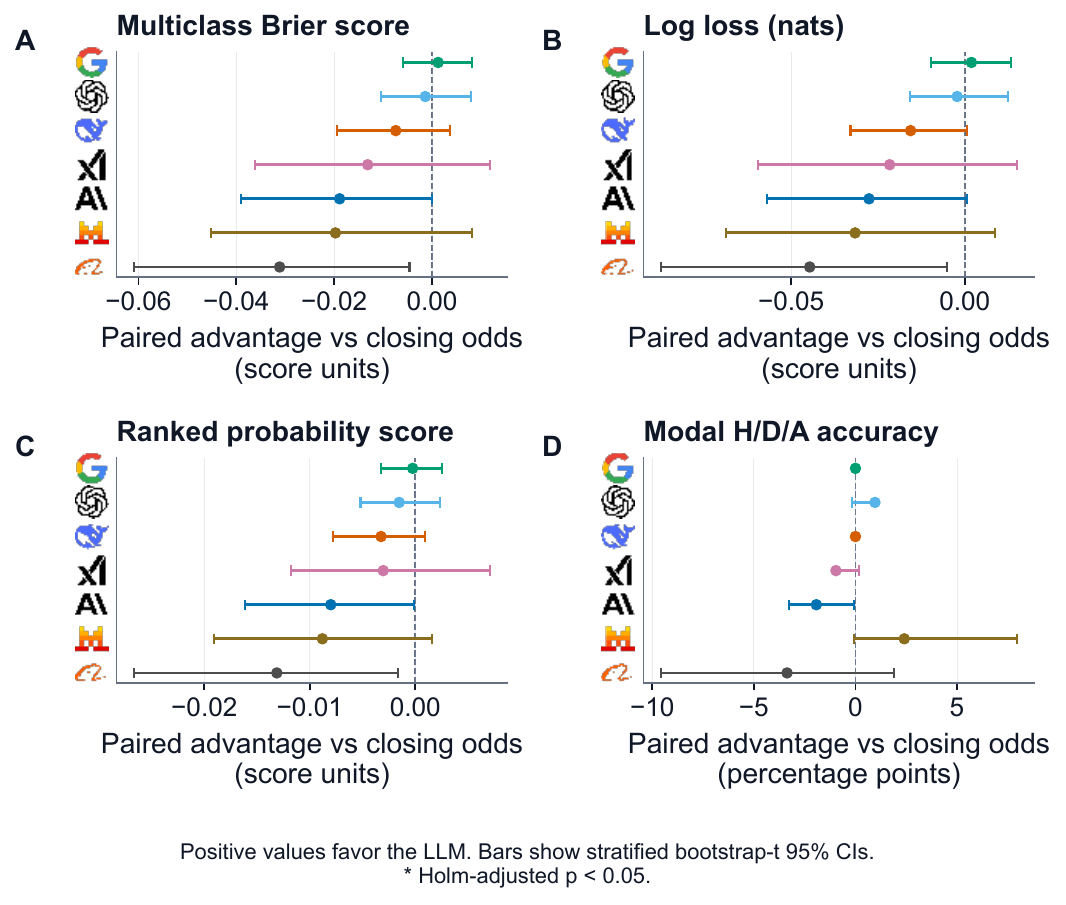}
  \caption{Paired LLM advantage over closing odds for Brier score, log loss, ranked probability score, and modal H/D/A accuracy. Positive values favor the LLM; bars are stratified bootstrap-$t$ $95\%$ intervals with inference adjusted across model versions within each metric.}
  \label{fig:closing-odds-paired}
\end{figure}

\begin{figure}[H]
  \centering
  \includegraphics[width=\textwidth,height=0.78\textheight,keepaspectratio]{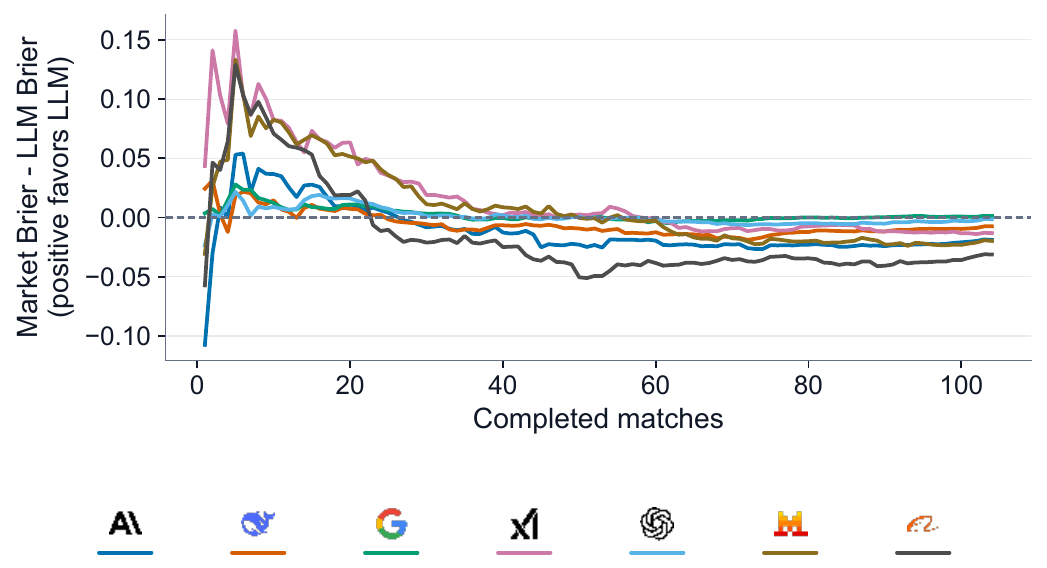}
  \caption{Cumulative Brier advantage relative to closing odds. Positive values indicate a lower cumulative mean Brier score for the LLM in the matched open-book, probabilistic forecast T--2h condition. Early trajectories are volatile and are interpreted descriptively.}
  \label{fig:closing-odds-cumulative}
\end{figure}

\begin{multicols}{2}
\subsection{External Baseline Eligibility}

External baselines enter the analysis only after a common provenance, timing, target-compatibility, and reproduction review. Timestamped bookmaker-level odds pass this review for all 104 matches. Opta and five coauthor-supplied classical football models were also considered: time-weighted ability Poisson, ordered-logit Elo, Elo--Poisson, Dixon--Coles, and ability Poisson with market value. They remain excluded because row-level provenance, exact timing, or faithful reproduction is unavailable.

We remove bookmaker overround before consensus aggregation. Closing odds appear only in this dedicated appendix comparison and are matched to the registered open-book, probabilistic forecast T--2h condition. Coauthor-supplied predictions are never used directly. A classical baseline can enter only after its supplied code is executed in an isolated pinned environment and all successfully reproduced candidates are retained without post hoc selection.
\end{multicols}

\twocolumn
\flushbottom

\clearpage
\section{Exact LLM Prompt Templates}
\label{app:prompts}

\begin{sloppypar}
This section reproduces the prompt builders used by the released project. Match prompts use template identifier \texttt{wc2026\_v1}; tournament-question prompts use \texttt{wc2026\_special\_v1}. The blocks follow the exact concatenation order in \path{packages/llm/src/prompt.ts} and \path{packages/llm/src/special-questions.ts}. Text in angle brackets denotes a runtime value; optional lines are emitted only for the relevant question type. The four benchmark variants combine one information-access header with one prompting-strategy instruction while holding the remaining blocks fixed.
\end{sloppypar}

\subsection{Tournament Questions and Candidate Lists}

The 15 questions are: \texttt{group\_winner\_A} through \texttt{group\_winner\_L}, \texttt{semifinalists}, \texttt{world\_champion}, and \texttt{top\_scorer\_team}. Group-winner questions are one-team questions over the four teams in the relevant group. The champion and top-scorer-team questions are one-team questions over all 48 teams. The semifinalist question is a fixed-set question with exactly four final selections and marginal inclusion probabilities for all 48 teams.

\begin{table*}[t]
  \centering
  \caption{Exact candidate teams supplied to the tournament-question prompts. The union of the 12 groups forms the 48-team candidate list for the semifinalist, champion, and top-scorer-team questions.}
  \label{tab:tournament-candidates}
  \small
  \begin{tabularx}{\textwidth}{@{}cX@{}}
    \toprule
    \textbf{Group} & \textbf{Candidate teams} \\
    \midrule
    A & Mexico; South Africa; South Korea; Czechia \\
    B & Canada; Switzerland; Qatar; Bosnia-Herzegovina \\
    C & Brazil; Morocco; Haiti; Scotland \\
    D & United States; Paraguay; Australia; Turkey \\
    E & Germany; Ecuador; Ivory Coast; Curaçao \\
    F & Netherlands; Japan; Tunisia; Sweden \\
    G & Belgium; Egypt; Iran; New Zealand \\
    H & Spain; Uruguay; Cape Verde Islands; Saudi Arabia \\
    I & France; Senegal; Iraq; Norway \\
    J & Argentina; Algeria; Austria; Jordan \\
    K & Portugal; Colombia; Uzbekistan; Congo DR \\
    L & England; Croatia; Ghana; Panama \\
    \bottomrule
  \end{tabularx}
\end{table*}

\subsection{Prompting Consistency}
Probabilistic forecast prompting also reduces agreement between the predicted scoreline and the most probable H/D/A outcome (e.g., $2{:}1$ with home win). Agreement decreases from $69.0\%$ to $62.4\%$ in the closed-book condition and from $83.8\%$ to $79.7\%$ in the open-book condition. Disagreement does not necessarily make a forecast invalid because one exact scoreline can be most likely even when another outcome has the highest total probability. Figures~\ref{fig:prompting-strategy} and~\ref{fig:prompting-agreement} in Appendix~\ref{app:prompting-details} summarize the accuracy, reported forecast, and agreement effects.

\subsection{Match-Level Forecast Prompt (\texttt{wc2026\_v1})}

The prompt concatenates the common preamble, one information-access header, one prompting-strategy instruction, the common definitions, the runtime fixture block, and the response schema.

\subsubsection*{Common preamble}
\promptheadingbreak
\begin{promptbox}{Common match prompt preamble}
You are forecasting a football match before it is played.
Return only valid JSON. Do not include markdown or any text outside JSON.
Return a concise reason only; do not reveal hidden reasoning or chain-of-thought.
Use calibrated probabilities and do not overstate certainty.
\end{promptbox}

\subsubsection*{Closed-book access header}
\promptheadingbreak
\begin{promptbox}{Closed-book prompt component}
Access condition: CLOSED_BOOK.
Do not use internet search, browsing, tools, APIs, external data sources, or project databases.
Use only the match information below plus your internal football knowledge.
You do not have access to this project's stored predictions, analytics, scores, or tournament-tree outputs.
\end{promptbox}

\subsubsection*{Open-book access header}
\promptheadingbreak
\begin{promptbox}{Open-book prompt component}
Access condition: OPEN_BOOK.
You may use the available web-search tool for current public factual information about teams, squads, injuries, form, fixtures, and tournament context.
Do not read, request, infer from, or use this project's stored predictions, analytics, scores, or tournament-tree outputs.
Base the final forecast on public information, the match information below, and calibrated football reasoning.
\end{promptbox}

\subsubsection*{Score-first instruction}
\promptheadingbreak
\begin{promptbox}{Score-first prompt component}
Prompt strategy: DIRECT_SCORE.
Choose the single most likely 90-minute scoreline first.
Then assign calibrated 90-minute home/draw/away probabilities, expected goals, full-match probabilities, and advancement probabilities consistent with that forecast.
Do not overstate certainty.
\end{promptbox}

\subsubsection*{Probabilistic forecast instruction}
\promptheadingbreak
\begin{promptbox}{Probabilistic forecast prompt component}
Prompt strategy: PROBABILISTIC_FORECAST.
Estimate calibrated 90-minute home/draw/away probabilities and expected goals first.
Then derive the single most likely 90-minute scoreline, full-match probabilities, and advancement probabilities from that forecast.
Do not overstate certainty.
\end{promptbox}

\subsubsection*{Common definitions block}
\promptheadingbreak
\begin{promptbox}{Common match definitions}
Definitions:
- 90-minute result means regulation time plus stoppage time, excluding extra time and penalties.
- home_win_90_prob is the probability that the listed home team leads after 90 minutes plus stoppage time.
- draw_90_prob is the probability that the match is tied after 90 minutes plus stoppage time.
- away_win_90_prob is the probability that the listed away team leads after 90 minutes plus stoppage time.
- expected_home_goals_90 and expected_away_goals_90 are expected goals scored in regulation time plus stoppage time.
- most_likely_score_90 is the single most likely score after regulation time plus stoppage time.
- Full-match result means final match outcome after all applicable extra time and penalty shootout procedures.
- For group-stage matches, full-match result is the same as the 90-minute result.
- For knockout matches, home_advances_prob and away_advances_prob are probabilities that each team advances/wins the tie after extra time and penalties if needed.
- Probabilities must be numbers between 0 and 1.
- home_win_90_prob + draw_90_prob + away_win_90_prob must sum to 1.
- home_win_full_prob + draw_full_prob + away_win_full_prob must sum to 1.
- For knockout matches, home_advances_prob + away_advances_prob must sum to 1.
- Confidence is the model's confidence in the overall forecast, between 0 and 1.
\end{promptbox}

The code-level field name \texttt{expected\_goals} denotes a predicted mean goal count, not the football expected-goals event statistic xG.

\subsubsection*{Runtime fixture block}
\promptheadingbreak
\begin{promptbox}{Runtime fixture template}
Match:
Sport: football / soccer
Competition: <COMPETITION>
Tournament edition: <TOURNAMENT_EDITION>
Stage: <STAGE>
Date UTC: <UTC_DATE>
Home/listed first team: <HOME_TEAM>
Away/listed second team: <AWAY_TEAM>
Venue: <VENUE_OR_UNKNOWN>
Is knockout match: <YES_NO_OR_UNKNOWN>
\end{promptbox}

\subsubsection*{Match-response schema}
\promptheadingbreak
\begin{promptbox}{Required JSON schema}
Return only valid JSON. Do not include markdown or explanation outside JSON.

JSON schema:
{
  "home_win_90_prob": number,
  "draw_90_prob": number,
  "away_win_90_prob": number,
  "expected_home_goals_90": number,
  "expected_away_goals_90": number,
  "most_likely_score_90": {
    "home": number,
    "away": number
  },
  "home_win_full_prob": number,
  "draw_full_prob": number,
  "away_win_full_prob": number,
  "most_likely_score_full": {
    "home": number,
    "away": number
  },
  "home_advances_prob": number or null,
  "away_advances_prob": number or null,
  "confidence": number,
  "reason": "short reason"
}
\end{promptbox}

\subsection{Tournament-Question Prompt (\texttt{wc2026\_special\_v1})}
\label{appendix:prompts}

All 15 tournament questions use the common preamble below. It is followed by one information-access header, one prompting-strategy instruction, a question-specific block, the complete candidate list, the static group and fixture context, and the relevant JSON schema.

\subsubsection*{Common preamble}
\promptheadingbreak
\begin{promptbox}{Common tournament prompt preamble}
You are forecasting \emph{2026 FIFA World Cup} tournament outcomes for Scoring System-style special questions.
Return only valid JSON. Do not include markdown or any text outside JSON.
Return concise reasoning_summary only; do not reveal hidden reasoning or chain-of-thought.
Use calibrated probabilities and valid candidate teams only.
These special predictions are one-time pre-tournament/STAGE_OPENING forecasts.
\end{promptbox}

\subsubsection*{Closed-book access header}
\promptheadingbreak
\begin{promptbox}{Closed-book prompt component}
Access condition: CLOSED_BOOK.
Do not use internet search, browsing, tools, APIs, external data sources, or project databases.
Use only the static tournament context below plus your internal football knowledge.
You do not have access to this project's stored match predictions, special predictions, analytics, scores, or tournament-tree outputs.
\end{promptbox}

\subsubsection*{Open-book access header}
\promptheadingbreak
\begin{promptbox}{Open-book prompt component}
Access condition: OPEN_BOOK.
You may use the available web-search tool for current public factual information about teams, squads, injuries, form, fixtures, and tournament context.
Do not read, request, infer from, or use this project's stored match predictions, special predictions, analytics, scores, or tournament-tree outputs.
Base the final forecast on public information, the static tournament context below, and calibrated football reasoning.
\end{promptbox}

\subsubsection*{Score-first instruction}
\promptheadingbreak
\begin{promptbox}{Score-first prompt component}
Prompt strategy: DIRECT_SCORE.
Make the final pick(s) first, then assign calibrated probabilities that support those pick(s).
Do not overstate certainty.
\end{promptbox}

\subsubsection*{Probabilistic forecast instruction}
\promptheadingbreak
\begin{promptbox}{Probabilistic forecast prompt component}
Prompt strategy: PROBABILISTIC_FORECAST.
Estimate calibrated probabilities for every valid candidate first, then derive the final pick(s).
Do not overstate certainty.
\end{promptbox}

\subsubsection*{Runtime question, candidate, and context blocks}
\promptheadingbreak
\begin{promptbox}{Runtime question and context template}
Special question:
question_id: <QUESTION_ID>
Question label: <QUESTION_LABEL>
Meaning: <QUESTION_MEANING>
prediction_type: <SINGLE_CHOICE_OR_MULTI_CHOICE_FIXED_K>
Use team names exactly as listed in the valid candidates section. Do not translate, abbreviate, or rename teams.
<OPTIONAL: Group: GROUP_NAME>
<OPTIONAL: Required number of final picks: K>
<TOP_SCORER_ONLY: Important: answer with the TEAM of the player who becomes top goalscorer, not the player name.>

Valid candidates:
- <EXACT_CANDIDATE_TEAM_NAME>
- <...>

Allowed static context:
Tournament edition: <TOURNAMENT_EDITION>

Groups:
Group <GROUP_NAME>: <TEAM_1>, <TEAM_2>, <TEAM_3>, <TEAM_4>
<...>

Official fixture data:
<UTC_DATE> | <STAGE> | <OPTIONAL_GROUP> | <HOME_TEAM> vs <AWAY_TEAM> | <OPTIONAL_VENUE>
<...>
\end{promptbox}

The instantiated question identifiers are \texttt{top\_scorer\_team}, \texttt{semifinalists}, \texttt{group\_winner\_A} through \texttt{group\_winner\_L}, and \texttt{world\_champion}.

\subsubsection*{one-team response schema}
\promptheadingbreak
\begin{promptbox}{Required one-team JSON schema}
Required JSON schema:
{
  "question_id": "<QUESTION_ID>",
  "prediction_type": "single_choice",
  "stage": "STAGE_OPENING",
  "choices": [
    {
      "team": "exact candidate team name",
      "probability": number,
      "rank": number,
      "is_final_pick": boolean
    }
  ],
  "final_pick": "exact candidate team name",
  "confidence": number,
  "reasoning_summary": "brief explanation"
}
Include exactly one choices entry for every valid candidate. For single_choice, probabilities must sum to 1.
Rank may be omitted or approximate; stored ranks are recalculated from probabilities during validation.
\end{promptbox}

\subsubsection*{Four-semifinalist response schema}
\promptheadingbreak
\begin{promptbox}{Required four-semifinalist JSON schema}
Required JSON schema:
{
  "question_id": "semifinalists",
  "prediction_type": "multi_choice_fixed_k",
  "k": 4,
  "stage": "STAGE_OPENING",
  "choices": [
    {
      "team": "exact candidate team name",
      "probability": number,
      "rank": number,
      "is_final_pick": boolean
    }
  ],
  "final_picks": ["exact candidate team name", "..."],
  "reasoning_summary": "brief explanation"
}
Include exactly one choices entry for every valid candidate. final_picks must contain exactly 4 unique teams.
Rank may be omitted or approximate; stored ranks are recalculated from probabilities during validation.
\end{promptbox}

\subsection{Single Schema-Repair Prompts}

A response that fails parsing or validation receives at most one repair call. Repair calls preserve the substantive forecast whenever possible and are stored separately from the original response.

\subsubsection*{Match-level repair prompt}
\promptheadingbreak
\begin{promptbox}{Match-level repair prompt}
Your previous response could not be parsed or validated.
Convert it into valid JSON matching the required schema.
Do not change the substantive forecast unless required to satisfy probability constraints.
Return only valid JSON. Do not include markdown or explanation outside JSON.

Validation errors:
- <VALIDATION_ERROR>

Required schema:
<MATCH_RESPONSE_SCHEMA_ABOVE>

Previous response:
<ORIGINAL_RESPONSE>
\end{promptbox}

\subsubsection*{Tournament-question repair prompt}
\promptheadingbreak
\begin{promptbox}{Tournament-question repair prompt}
Your previous response could not be parsed or validated.
Convert it into valid JSON matching the required special-question schema.
Do not change the substantive forecast unless required to satisfy candidate, probability, or pick constraints.
Return only valid JSON. Do not include markdown or explanation outside JSON.

Validation errors:
- <VALIDATION_ERROR>

<QUESTION_BLOCK>

<CANDIDATE_BLOCK>

<QUESTION-SPECIFIC_RESPONSE_SCHEMA_ABOVE>

Previous response:
<ORIGINAL_RESPONSE>
\end{promptbox}

\end{document}